\documentclass{article} %
\usepackage{arxiv}

\usepackage{amsmath,amsfonts,bm}

\def\eqref#1{equation~\ref{#1}}
\def\1{\bm{1}}

\newcommand{\wrt}{w.r.t.\ }

\def\eps{{\epsilon}}
\def\btt{{\boldsymbol{\theta}}}

\def\rg{{\textnormal{g}}}

\def\ry{{\textnormal{y}}}

\def\rvx{{\mathbf{x}}}

\def\rvz{{\mathbf{z}}}

\def\ervg{{\textnormal{g}}}

\def\ervy{{\textnormal{y}}}

\def\vf{{\bm{f}}}

\def\vw{{\bm{w}}}
\def\vx{{\bm{x}}}

\DeclareMathAlphabet{\mathsfit}{\encodingdefault}{\sfdefault}{m}{sl}
\SetMathAlphabet{\mathsfit}{bold}{\encodingdefault}{\sfdefault}{bx}{n}

\def\gD{{\mathcal{D}}}

\def\gG{{\mathcal{G}}}
\def\gH{{\mathcal{H}}}

\def\gX{{\mathcal{X}}}
\def\gY{{\mathcal{Y}}}
\def\gZ{{\mathcal{Z}}}

\def\sP{{\mathbb{P}}}

\def\sR{{\mathbb{R}}}

\newcommand{\sigmoid}{\sigma}

\newcommand{\norm}[1]{\left\lVert#1\right\rVert}
\newcommand{\set}[1]{\left\{#1\right\}}

\usepackage{hyperref}
\usepackage{url}
\usepackage[capitalize,noabbrev]{cleveref}

\hypersetup{
    colorlinks,
    linkcolor={red!50!black},
    citecolor={blue!50!black},
    urlcolor={blue!80!black}
}

\usepackage{algorithm}
\usepackage{subcaption}
\usepackage[ruled,vlined,algo2e]{algorithm2e}
\usepackage[utf8]{inputenc} % allow utf-8 input
\usepackage[T1]{fontenc}    % use 8-bit T1 fonts
\usepackage{booktabs}       % professional-quality tables
\usepackage{amsfonts}       % blackboard math symbols
\usepackage{nicefrac}       % compact symbols for 1/2, etc.
\usepackage{microtype}      % microtypography
\usepackage{xcolor} % colors
\usepackage{amsmath}
\usepackage{amssymb}
\usepackage{mathtools}
\usepackage{amsthm}
\usepackage{wrapfig}
\usepackage{bm}
\usepackage{stmaryrd}
\usepackage{array}
\usepackage{tabularx} % in the preamble
\usepackage{multirow}
\usepackage{diagbox}
\usepackage{autonum}
\usepackage{enumitem}
\usepackage{natbib}
  \setlist{leftmargin=*}
  
\usepackage{pifont}% http://ctan.org/pkg/pifont
\newcommand{\cmark}{\ding{51}}%
\newcommand{\xmark}{\ding{55}}%

\usepackage{todonotes} % for comments

\newcommand{\grasp}{\textsc{GraSP}}

\setlist[itemize]{noitemsep, nolistsep}
\setlist[enumerate]{noitemsep, nolistsep}

\title{Outlier-Robust Group Inference via Gradient Space Clustering}
\author{Yuchen Zeng \thanks{Work performed while doing an internship at IBM Research.} \\
Department of Computer Science\\
University of Wisconsin-Madison\\
\texttt{yzeng58@wisc.edu} \\
\And
Kristjan Greenewald \\
IBM Research \\
MIT-IBM Watson AI Lab \\
\texttt{kristjan.h.greenewald@ibm.com} \\
\AND
Kangwook Lee \\
Department of Electrical and Computer Engineering\\
University of Wisconsin-Madison\\
\texttt{kangwook.lee@wisc.edu} \\
\AND
Justin Solomon \\
Department of Electrical Engineering \& Computer Science\\
Massachusetts Institute of Technology\\
\texttt{jsolomon@mit.edu} \\
\And
Mikhail Yurochkin \\
IBM Research \\
MIT-IBM Watson AI Lab \\
\texttt{mikhail.yurochkin@ibm.com} \\
}
\date{}

\begin{document}

\maketitle
\begin{abstract}

Traditional machine learning models focus on achieving good performance on the overall training distribution, but they often underperform on minority groups.
Existing methods can improve the worst-group performance, but they can have several limitations: (i) they require group annotations, which are often expensive and sometimes infeasible to obtain, and/or (ii) they are sensitive to outliers. Most related works fail to solve these two issues simultaneously as they focus on conflicting perspectives of minority groups and outliers. We 
address the problem of learning group annotations in the presence of outliers by clustering the data in the space of gradients of the model parameters. We show that data in the gradient space has a simpler structure while preserving information about minority groups and outliers, making it suitable for standard clustering methods like DBSCAN. Extensive experiments demonstrate that our method significantly outperforms state-of-the-art both in terms of group identification and downstream worst-group performance.
% \my{we can cut sentence mentioning related work}

% Traditional machine learning algorithms focus on achieving good performance on the overall training distribution, which may result in low accuracy on certain subgroups, especially when subgroups are highly unbalanced. 
% Distributionally Robust Optimization (DRO) is a method for improving the worst-subgroup performance, which is equivalent to reweighting the samples from different subgroups. 
% However, DRO suffers from two serious issues: i) requires subgroup annotations, and ii) vulnerable to outliers. 
% Unfortunately, most of the existing approaches fail to solve the two issues simultaneously, and the methods handle each of the issues often conflict with each other.  
% In this work, we propose Gradient Subgroup Scanning (GraSS), which perform clustering on the gradient of samples for both identifying subgroup annotations and outliers.
% Via extensive experiments, we show that GraSS significantly outperforms state-of-the-art methods in terms of both subgroup annotation prediction and worst-subgroup performance.
\end{abstract}

\section{Introduction}
\label{sec:intro}

% \begin{quote}
%     \textit{Happy families are all alike; every unhappy family is unhappy in its own way.}\\
%     \hspace*{\fill}- Anna Karenina\quad 
% \end{quote}

Empirical Risk Minimization (ERM), i.e., the minimization of average training loss over the set of model parameters, is the standard training procedure in machine learning. It yields models with strong in-distribution performance\footnote{I.e. low loss on test data drawn from the same distribution as the training dataset.} but does not guarantee satisfactory performance on minority groups that contribute relatively few data points to the training loss function \citep{sagawa2019distributionally,koh2021wilds}. 
% due to their relatively low contribution to the loss. 
This effect is particularly problematic when the minority groups correspond to socially-protected groups. For example, in the toxic text classification task, certain identities are overwhelmingly abused in online conversations that form data for training models detecting toxicity \citep{dixon2018measuring}. 
% \justin{training data for what?}
Such data lacks sufficient non-toxic examples mentioning these identities, yielding
problematic and unfair spurious correlations -- as a result ERM learns to associate these identities with toxicity \citep{dixon2018measuring,garg2019counterfactual,yurochkin2020sensei}. A related phenomenon is \emph{subpopulation shift} \citep{koh2021wilds}, i.e., when the test distribution differs from the train distribution 
%\justin{is that really what's going on here? it seems like your test data could also have the same lack of minority representation but you still want to be fair to the minority}\my{that is the definition of subpopulation shift - it is not always related to fairness, could be just test performance under distribution shift. I edited to clarify} 
in terms of group proportions. Under subpopulation shift, poor performance on the minority groups in the train data translates into poor overall test distribution performance, where these groups are more prevalent or more heavily weighted. Subpopulation shift occurs in many application domains \citep{tatman2017gender,beery2018recognition,oakden2020hidden,santurkar2020breeds,koh2021wilds}.
% \my{we need some example of subpopulation shift if not cows (it is one used in some DRO papers). We can also just name a few domains where this happens with citations}% For example, train data may have majority of cows appearing on a grassy background resulting in a model with poor performance on a dataset with cows on other backgrounds \my{cite}.%<--- doesn't add anything

Prior work offers a variety of methods for training models robust to subpopulation shift and spurious correlations, including group distributionally robust optimization (gDRO) \citep{hu2018does,sagawa2019distributionally}, importance weighting \citep{shimodaira2000improving,byrd2019effect}, subsampling \citep{sagawa2020investigation,idrissi2022balancing,maity2022does}, and variations of tilted ERM \citep{li2020tilted,li2021tilted}. These methods are successful in achieving comparable performance across groups in the data, but they require group annotations. The annotations can be expensive to obtain, e.g., labeling spurious backgrounds in image recognition \citep{beery2018recognition} or labeling identity mentions
% \kl{mentioned?}
in the toxicity example. It also could be challenging to anticipate all potential spurious correlations in advance, e.g., it could be background, time of day, camera angle, or unanticipated identities subject to harassment.

% emerged as method for achieving comparable performance across groups, even those underrepresented in the data. They need additional labeling which is also ambiguous as we may not know the soruce of spuriosu correlation. It could be background, specific identity words, camera angle, time of day, etc.

Recently, methods have emerged for learning group annotations \citep{sohoni2020george,liu2021jtt,creager2021eiil} and variations of DRO that do not require groups \citep{hashimoto2018fairness,zhai2021doro}. One common theme is to treat data where an ERM model makes mistakes (i.e., high-loss points) as a minority group \citep{hashimoto2018fairness,liu2021jtt} and increase the weighting of these points. Unfortunately, such methods are at risk of overfitting to outliers (e.g., mislabeled data, corrupted images), which are also high-loss points. Indeed, existing methods for outlier-robust training propose to \emph{ignore} the high-loss points \citep{shen2019learning}, the direct opposite of the approach in \citep{hashimoto2018fairness,liu2021jtt}. 

\begin{figure}
    \centering
    \includegraphics[width=\textwidth]{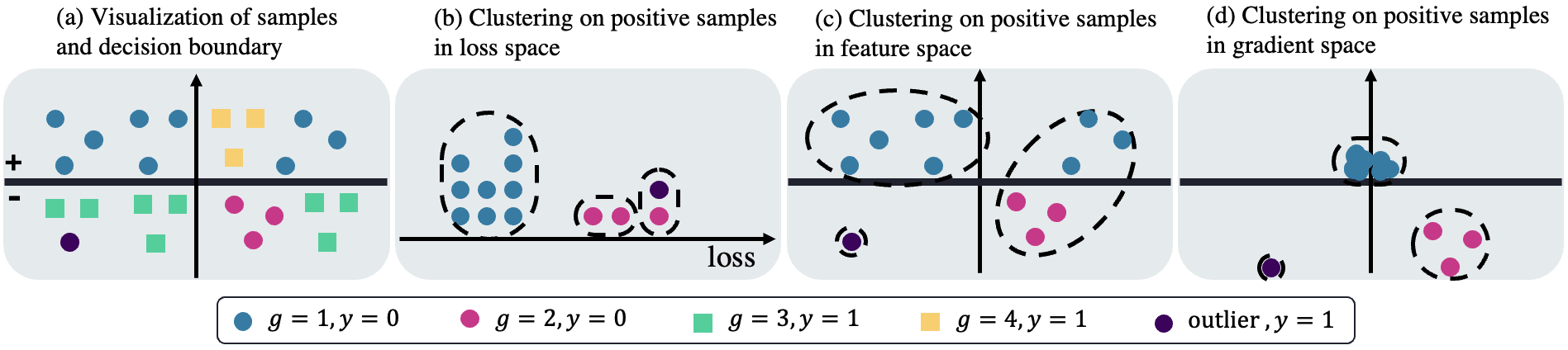}
    \caption{\textbf{An illustration of learning group annotations in the presence of outliers.} 
    % the failure of clustering in feature space and the benefit of clustering in gradient space.}
    (a) A toy dataset in two dimensions. There are four groups $g=1,2,3,4$ and an outlier. $g=1$ and $g=3$ are the majority groups distributed as mixtures of three components each; $g=2$ and $g=4$ are unimodal minority groups. $y$-axis is the decision boundary of a logistic regression classifier. Figures (b, c, d) compare different data views for learning group annotations and detecting outliers via clustering of samples with $y=0$. (b) loss values can confuse outliers and minority samples which both can have high loss; (c) in the original feature space it is difficult to distinguish one of the majority group modes and the minority group; (d) gradient space (bias gradient omitted for visualization) simplifies the data structure making it easier to identify the minority group and to detect outliers. 
    % \kl{update figure (a) so that the blue points are vertically algined? 4-2-3?}
    % $\hat{y} = \sI\set{\sigma(\vw^\top \vx + b) > .5}$. \yuchen{More description...}
    % Group inference via clustering is performed conditional on class. \yuchen{Is ''conditional on class'' a right expression?}
    % This example focuses on positive class only.
    % The arrows in (c) depict the gradient of data points .
    % The dashed circles in (b) and (d) show the resulted clusters led by the corresponding methods. 
    % We observe that 
    % \yuchen{This figure will be updated...}
    % \yuchen{Add the x,y labels?}
    }
    \label{fig:motivating}
\end{figure}

In this paper, our goal is to learn group annotations in the presence of outliers. Rather than using loss values (which above were seen to create opposing tradeoffs), we propose to instead first represent data using gradients of a datum's loss \wrt the model parameters. Such gradients tell us how a specific data point wants the parameters of the model to change to fit it better. In this gradient space, we anticipate groups (conditioned on label) to correspond to gradients forming clusters. Outliers, on the other hand, majorly correspond to isolated gradients: they are likely to want model parameters to change differently from any of the groups \emph{and} other outliers. See Figure \ref{fig:motivating} for an illustration.
% in other words, they are all ``unhappy in their own way.''
The gradient space structure allows us to separate out the outliers and learn the group annotations via traditional clustering techniques such as DBSCAN \citep{ester1996dbscan}. We use learned group annotations to train models with improved worst-group performance (measured \wrt the true group annotations). 
% \kl{Should we mention Figure 1 here?}

We summarize our contributions below:
\begin{itemize}
    \item We show that gradient space simplifies the data structure and makes it easier to learn group annotations via clustering.
    \item We propose Gradient Space Partitioning (\grasp), a method for learning group annotations in the presence of outliers for training models robust to subpopulation shift. 
    % The first step is to train an ERM model and perform data clustering in the corresponding gradient space using DBSCAN, and the second is to train a model with the learned groups using gDRO.
    % \item We provide intuition for the advantages of clustering in the gradient space 
    \item We conduct extensive experiments on one synthetic dataset and three datasets from different modalities and demonstrate that our method achieves state-of-the-art both in terms of group identification quality and downstream worst-group performance.
\end{itemize}

% Gradient space also simplifies the groups structure, which otherwise might have complicated densities. EXAMPLE or don't say this? Contributions?

\section{Preliminaries and Related Work}

% In this section, we review the subpopulation shift setting \citep{koh2021wilds} and describe the problem we address in this work.
In this section, we review the problem of training models in the presence of minority groups.
Denote $[N] = \set{1,\dots,N}$.
Consider a dataset $\gD = \set{\rvz}_{i=1}^{n}  \subset \gZ$ consisting of $n$ samples $\rvz \in \gZ$, $\rvz = (\rvx, \ervy)$, where $\rvx \in \gX = \sR^d$ is the input feature and $\ervy \in \gY = \{1,\ldots,C\}$ is the class label. 
% Denote the set of samples with $\ry = y$ as $\gG^{(y)}$, $k\in \gY$.
The samples from each class $y\in\gY$ are categorized into $K_y$ groups.
% \justin{which may not be known at training time?}\yuchen{For Group-aware methods, the group partition is known. We describe these settings below}. 
% For example, in the context of fair learning, these groups can correspond to differences in gender and race; or perhaps differences in geographic and linguistic background.
Denote $K$ to be the total number of %pre-defined 
groups $\set{\gG_1, \dots, \gG_K} \triangleq P \subset \gZ$, where $K = \sum_{y\in\gY} K_y$. 
Denote the group membership of each point in the dataset as $\set{\rg_i}_{i=1}^{n}$, where $\rg_i \in [K]$ for all $i \in [n]$. For example, in toxicity classification, a group could correspond to a toxic comment mentioning a specific identity, or, in image recognition, a group could be an animal species appearing on an atypical background \citep{beery2018recognition,sagawa2019distributionally}.
% \kl{In the toy example, we have $g=-1$, so I am confused. Also, shouldn't there be some constraint on $g_i$ depending on the value of $y_i$?}

The goal is to to learn a model $h \in \gH: \gX \to \gY$ parameterized by $\btt \in \Theta$ that performs well on all groups $\gG_k$, where $k\in[K]$. Depending on the application, this model can alleviate fairness concerns \citep{dixon2018measuring}, remedy spurious correlations in the data \citep{sagawa2019distributionally}, and promote robustness to subpopulation shift \citep{koh2021wilds}, i.e., when the test data has unknown group proportions. 
% This sentence is the same as the first sentence of this paragraph --- removed by Yuchen ---
% Concretely, the goal is to learn a model $h \in \gH: \gX \to \gY$ parameterized by $\btt \in \Theta$ that performs well on all groups $\gG_k$, where $k\in[K]$.

% Note that the ``subpopulation shift" terminology is due to the fact that if the model performs well on all groups, the model will be robust to ``shifts" in the proportions of membership in the various groups, i.e. if the test dataset has different group membership proportions than the train dataset. 

We divide the approaches for learning in the presence of minority groups into three categories: the \emph{group-aware} setting where the group annotations $\rg_i$ are known, the \emph{group-oblivious} setting that does not use the group annotations, and the \emph{group-learning} setting where the group annotations are learned from data to be used as inputs to the group-aware methods.

\textbf{Group-aware setting.} 
% The \emph{group-aware} setting considers the cases that group annotations of the data samples are available. 
% In fact, m
Many prior works %study the problem of learning in the presence of minority groups assuming 
assume access to the minority group annotations. Among the state-of-the-art methods in this setting is group Distributionally Robust Optimization (gDRO) \citep{sagawa2019distributionally}. 
Let $\ell: \gY \times \gY$ be a loss function. 
The optimization problem of gDRO is 
\begin{equation}\label{opt:gdro}
    \min_{\btt\in\Theta} \max_{k\in[K]} \frac{1}{|\gG_{k}|}\sum_{\rvz \in \gG_{k}} \ell(\ervy, h_\btt(\rvx)), \tag{gDRO}
\end{equation}
which aims to minimize the maximum group loss. 
In addition to assuming clean group annotations, another line of research under this setting considers noisy or partially available group annotations~\citep{jung2022learning,lamy2019noise,mozannar2020fair,celis2021fair}. 
% Confidence-based Group Label (CGL)~\citep{jung2022learning} train a group classifier based on partially available group annotations.
% , and randomly assign group labels to the samples with low-confident group predictions
Methods in this class achieve meaningful improvements over ERM in terms of worst-group accuracy, but anticipating relevant minority groups and obtaining the annotations is often burdensome.% limiting their practical utility.
% The formulation of this optimization problem implies that applying gDRO requires group annotations. 

% \kl{Also, I think we should mention a line of work which considers noise protected group labels. }\my{we can add partially/noisy references to the group aware paragraph}

\textbf{Group-oblivious setting.} In contrast to the group-aware setting, the \emph{group-oblivious} setting attempts to improve worst-group performance without group annotations. Methods in this group rely on various forms of DRO \citep{hashimoto2018fairness,zhai2021doro} or adversarial reweighing \citep{lahoti2020fairnessarl}. Algorithmically, this results in up/down-weighing the contribution of the high/low-loss points. For example, \citet{hashimoto2018fairness} optimizes a DRO objective with respect to a chi-square divergence ball around the data distribution, which is equivalent to minimizing $\frac{1}{n}\sum_i [\ell(\ervy, h_\btt(\rvx)) - \eta]^2_+$, i.e., an ERM discounting low-loss points by a constant depending on the ball radius.

% considers the case when group annotations are unknown. 
% This is the focus of our work. 
% The most popular method for handling group-oblivious subpopulation shift problem is \emph{Distributionally Robust Optimization} (DRO).
% DRO improves the worst-group performance by upweighting the samples with high loss. 
% However, \citet{zhai2021doro} observed that DRO suffers from performance degradation and training instability in the presence of outliers, because the outliers generally have a high loss and are upweighted by DRO.
%The large weights attached to outliers make DRO vulnerable to outliers.

\textbf{Group-learning setting.}
% \kl{What about ``Fairness without Demographics through Adversarially Reweighted Learning''?}
% \kl{Also, semi supervised setting -- Learning Fair Classifiers with Partially Annotated Group Labels}
The final category corresponds to a two-step procedure, wherein the data points are first assigned group annotations based on various criteria, followed by group-aware training typically using gDRO. In this category, Just Train Twice (JTT) \citep{liu2021jtt} trains an ERM model and designates high-loss points as the minority and low-loss points as the majority group;
% Adversarially Reweighted Learning (ARL)~\citep{lahoti2020fairnessarl} softly assign group annotations to samples by maximizing a variant of gDRO objective, \yuchen{Add to Table 1?}
% assumes that the group annotations are partially available, and train a group classifier based on that
George \citep{sohoni2020george} seeks to cluster the data to identify groups with a combination of dimensionality reduction, overclustering, and augmenting features with loss values, and Environment Inference for Invariant Learning (EIIL) \citep{creager2021eiil} finds group partition that maximizes the Invariant Risk criterion \citep{arjovsky2019invariant}.

Our method, Gradient Space Partitioning (GraSP), belongs to this category. GraSP differs from prior works in its ability to account for outliers in the data. In addition, prior methods in this and the group-oblivious categories typically require validation data with \emph{true} group annotations for model selection to achieve meaningful worst-group performance improvements over ERM, while GraSP does not need these annotations to achieve good performance. In our experiments, this can be attributed to GraSP's better recovery of the true group annotations, making them suitable for gDRO model selection (see Section \ref{sec:exp}). We summarize properties of the most relevant methods in each setting in Table \ref{tab:baseline}. % \my{Yuchen, please move table here, remove FeaSP, and add JTT and Hashimoto et al}

% Describe JTT, George, EIIL; our method is in this group. Mention that they often rely on validation group info.
\begin{table}[t]
\caption{\textbf{Summary of methods for learning in the presence of minority groups. }
"-" indicates that there is no clear evidence in the prior works.
% Note that \grasp{} is the only method designed for achieving 
}
\label{tab:baseline}
% \vskip 0.15in
\begin{center}
\begin{small}\tiny{
\setlength{\tabcolsep}{1.2pt}
\begin{tabularx}{\textwidth}{p{8em}|c|c|cc|cccc}
\toprule
Setting & & Group-aware & \multicolumn{2}{c}{Group-oblivious} & \multicolumn{4}{c}{Group-learning} \\ \midrule
Method & \multirow{2}{*}{ERM} & gDRO & $\chi^2$-DRO & DORO & JTT & EIIL & George   & \grasp{} \\
& & \citep{sagawa2019distributionally} & \citep{hashimoto2018fairness} & \citep{zhai2021doro} & \citep{liu2021jtt} & \citep{creager2021eiil} & \citep{sohoni2020george} & (Ours) \\
\midrule
Improves worst-group performance? & \xmark& \cmark& \cmark& \cmark & \cmark & \cmark    & \cmark & \cmark \\ 
No training group annotations? & \cmark& \xmark & \cmark& \cmark& \cmark& \cmark & \cmark  & \cmark \\ % \midrule
No validation group annotations? & \cmark & \xmark& \xmark & \xmark& \xmark& \xmark & \cmark   & \cmark \\
Group inference? & \xmark & \xmark& \xmark & \xmark & \cmark & \cmark & \cmark & \cmark\\
Robust to outliers? & \xmark & - & \xmark & \xmark& \xmark & - & - & \cmark \\ % \midrule
\bottomrule
\end{tabularx}
}
\end{small}
\end{center}
\vskip -0.15in
\end{table}

\textbf{The challenge of outliers.}
% Outliers, e.g., mislabeled samples or corrupted images, are ubiquitous in applications \citep{singh2012outlier}, and training ML models robust to outliers has long been a topic of inquiry \my{CITE}. Outliers are especially challenging when training in the presence of (unknown) minority groups, which could be hard to distinguish from outliers but require the opposite treatment: 
Outliers, e.g., mislabeled samples or corrupted images, are ubiquitous in applications \citep{singh2012outlier}, and outlier detection has long been a topic of inquiry in ML \citep{hodge2004survey,wang2019progress}. Outliers are especially challenging to detect when data has (unknown) minority groups, which could be hard to distinguish from outliers but require the opposite treatment:
Minority groups need to be upweighted while outliers must be discarded. \citet{hashimoto2018fairness} write, ``it is an open question whether it is possible to design algorithms which are both fair to unknown latent groups and robust [to outliers].''

We provide an illustration of a dataset with minority groups and an outlier in Figure \ref{fig:motivating}(a). Figure \ref{fig:motivating}(b) illustrates the problem with the methods relying on the loss values. Specifically, \citet{liu2021jtt} and \citet{hashimoto2018fairness} upweigh high-loss points, overfitting the outlier. \citet{zhai2021doro} optimize \citet{hashimoto2018fairness}'s objective function after discarding a fraction of points with the largest loss values to account for outliers. They assume that outliers will have higher loss values than the minority group samples, which can easily be violated leading to exclusion of the minority samples, as illustrated in Figure~\ref{fig:motivating}.

% Explain what are outliers; they are easy to confuse with minority; and use fig1 to show how all previous methods fail.

\textbf{Gradients as data representations.}
Given a model $h_{\btt_0}(\cdot)$ and loss function $\ell(\cdot,\cdot)$,
% \footnote{The model and loss used here can differ from those used in downstream classification tasks.}
one can consider an alternative representation of the data where each sample is mapped to the gradient with respect to the model parameters of the loss on this sample:
% \kl{why arrow, not just equality? Also, explain what $\theta_0$ is here}
\begin{equation}
\label{eq:gradient}
    \vf_i = \left.\frac{\partial \ell (\ry_i, h_{\btt}(\rvx_i))}{\partial \btt} \right|_{\btt = \btt_0}\text{ for }i=1,\ldots,n.
\end{equation}
% where $\theta_0$ are some given model parameters (e.g., parameters of a trained ERM model). 
We refer to \eqref{eq:gradient} as the \emph{gradient representation}. Prior works considered gradient representations \citep{mirzasoleiman2020coresets}, as well as loss values \citep{shen2019learning}, for outlier-robust learning. Gradient representations have also found success in novelty detection~\citep{kwon2020novelty}, anomaly detection~\citep{kwon2020backpropagated}, and out-of-distribution inputs detection~\citep{huang2021importance}.

In this work, we show that, unlike loss values, gradient representations are suitable for simultaneously learning group annotations \emph{and} detecting outliers. Compared to the original feature space, gradient space simplifies the data structure, making it easier to identify minority groups. Figure \ref{fig:motivating}(c) illustrates a failure of feature space clustering. Here the majority group for class $y=0$ is a mixture of three components with one of the components being close to the minority group in the feature space. In the gradient space, for a logistic regression model, representations of misclassified points remain similar to the original features, while the representations of correctly classified points are pushed towards zero. We illustrate the benefits of the gradient representations in Figure \ref{fig:motivating}(d) and provide additional details in the subsequent section.

\section{\grasp{}: Gradient Space Partitioning}

% \begin{table}[t]
% \caption{Baselines. '-' indicates that there is no clear evidence. \yuchen{Change ``Unclear" to -?}}
% \label{tab:baseline}
% % \vskip 0.15in
% \begin{center}
% \begin{small}\tiny{

% \begin{tabularx}{.8\textwidth}{lccccccc}
% \toprule
% Method & ERM & EIIL & George & DORO & FeaSP & gDRO & \grasp{}(Ours) \\\midrule
% Distributionally robust? & \xmark & \cmark & \cmark & \cmark & \cmark & \cmark & \cmark \\ 
% Outlier robust? & \xmark & Unclear & Unclear & \cmark & Unclear & \xmark & \cmark \\ 
% Group inference? & \xmark & \cmark & \cmark & \cmark & \cmark & \xmark & \cmark \\
% No training group annotations? & \cmark & \cmark & \xmark & \cmark & \cmark & \xmark & \cmark \\
% No validation group annotations? & \cmark & \xmark & \cmark & \xmark & \cmark & \xmark & \cmark \\
% \bottomrule
% \end{tabularx}
% }
% \end{small}
% \end{center}
% \vskip -0.15in
% \end{table}

In this section, we present our method for group inference and outlier detection, which we refer to as Gradient Space Partitioning (\grasp{}).
We first demonstrate that the gradient space is more suitable for using clustering methods to learn group annotations and identify outliers than the feature space. 
We support this claim with an example using a logistic regression model and an empirical study of synthetic and semi-synthetic datasets.
% We first illustrate why gradient of a sample's loss \wrt the models parameters are more desirable representations for group inference with a motivating example.
% We then provide a theoretical interpretation of using such gradient-based representation for group inference based on logistic regression model. 
% Moreover, we numerically compare the clustering performance on the feature space and such gradient space to further verify that the gradient space is more suitable for learning group annotations via clustering than the feature space. 
We then present the details of \grasp{} in Sec.~\ref{sec:grasp_alg}. 

\subsection{Gradient Space vs Feature Space}\label{sec:space}

\textbf{Logistic regression example.}
We present an example based on the logistic regression model to better understand how using the gradient space simplifies the data structure and aids clustering.

Consider a binary classification problem ($\ry \in \{0,1\}$) and logistic regression model $\sP(\ry = 1|\rvx) = \sigma(\vw^\top \rvx + b)$ trained on the given dataset $\gD$, where $\sigma(\cdot)$ denotes the sigmoid function, $\vw$ are the coefficients and $b$ is the bias. Recall that the logistic regression loss is 
\[
\ell(\ry,\sigmoid(\vw^\top \rvx + b)) = -\ry \log(\sigmoid(\vw^\top \rvx + b)) - (1-\ry)\log(1-\sigmoid(\vw^\top \rvx + b)).
\]
The gradient of this loss at point $\rvz$ \wrt $(\vw, b)$ is 
\begin{equation}
\vf =: \nabla_{[\vw,b]}\ell(\ry,\vw^\top \rvx + b) = (\sigmoid(\vw^\top \rvx + b) - \ry)  \left[\begin{array}{c}\rvx\\ 1\end{array}\right].
\label{eq:grad}
\end{equation}
% \justin{what's the difference between $\ry$ and $y$?}.
%Let $\rvx = x \rvu$, where $x = \norm{\rvx}_2$; here, $\rvu = \nicefrac{\rvx}{\norm{\rvx}_2}$ is the unit vector with the same direction as $\rvx$. 
%The gradient $\vf$ can be written as a function of $x$ if we fix the direction $\rvu$, \ie, $\vf=(\sigma(x \vw^\top\vu)-\ry)(x\rvu, 1)$.
Note that this gradient is simply a scaling of the data vector $\rvx$ by the error $(\sigmoid(\vw^\top \rvx + b) - \ry) \in [-1,1]$, padded by an additional element (the bias entry) consisting of the error alone. In particular, note that when $\rvz$ is correctly classified, the scaling $(\sigma( \vw^\top\rvx + b)-\ry)$ will be close to zero, and when it is incorrectly classified, the magnitude of the scaling will approach 1. 

%To see this, we visualize how $(\sigma( \vw^\top\rvx)-\ry)x$ varies \wrt $x$ when $\vw^\top \rvu = 1$, $\ry=1$ in Fig.~\ref{fig:theo} in the supplement. 
\begin{wrapfigure}[25]{r}{0.3\textwidth}
    \centering
    \vspace{-.23in}
    \includegraphics[width=0.3\textwidth]{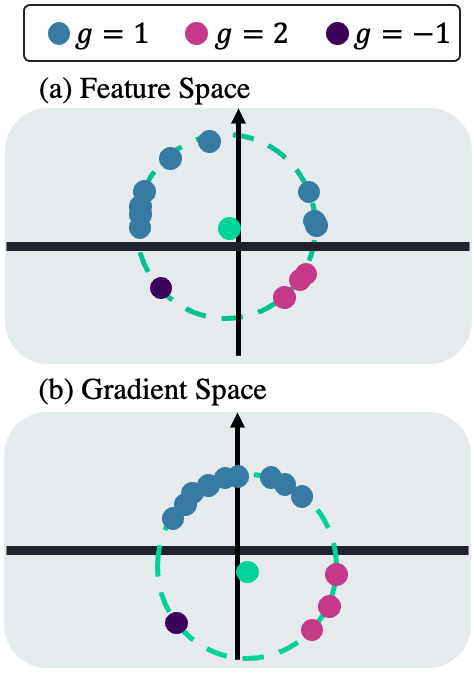}
    \vspace{-.2in}
    \caption{
    \textbf{
    % Cosine distance between the points from group 1 \wrt feature space and gradient space
    Normalized representations of the data from Figure~\ref{fig:motivating}(a) in (a) feature space and (b) gradient space.
    %\kl{Why is this cosine similarity?}
    }
    The green points are the means (before normalization) of the corresponding representations. Gradient space makes it easier to identify groups and detect outliers via clustering with centered cosine distances.
    % We observe that in the gradient space, minority gr, majority, and outlier it is easier to identify groups andpoints from the same group are clustered together to form high-density regions, while the same phenomenon is not observed in the feature space.
    }
    \vspace{-.3in}
    \label{fig:cosine}
\end{wrapfigure}

We interpret this gradient (\eqref{eq:grad}) through the lens of Euclidean distance ($\|\vf_i - \vf_j\|_2$) and centered cosine distance ($1 - \frac{\langle\vf_i - \mu_f, \vf_j - \mu_f \rangle}{\|\vf_i - \mu_f\|_2 \cdot \|\vf_j - \mu_f\|_2}$) metrics,\footnote{Here $\mu_f$ refers to the class-conditional empirical mean of $\vf$.}
% \kl{In the equation at the bottom of p5, it seems to be the mean of f across all classes}}
respectively. Recall that we apply clustering to each class independently.

\begin{itemize}
\item \textbf{Euclidean distance.} %\justin{this sentence feels clunky but I didn't know how to fix}
The scaling effect mentioned in the previous paragraph shrinks the correctly classified points towards the origin, while leaving the misclassified points almost unaffected. The error itself is included as an extra element (using loss as an additional feature was previously considered as a heuristic in feature clustering for learning group annotations \citep{sohoni2020george}).
Consequently, gradient clustering \wrt Euclidean distance should cluster the correctly classified samples into one ``majority'' group, and then divide the remaining points into minority groups and outliers based on the size of the error and their position in the feature space. For a visual example, see Figure~\ref{fig:motivating}(d).

\item \textbf{Centered cosine distance.}
% For convenience, define $\rvx' = \left[\begin{array}{c}\rvx\\ 1\end{array}\right]$. 
We compare the (class conditioned; class dependency omitted for simplicity) centering terms in the gradient and feature spaces, $\mu_f$ and $\mu_{x}$ respectively: 
% \kl{Why not just say $\mu_f$ is the mean of $f$ and $\mu_x$ is the mean of $x$? seems redundant.}
\[
\mu_f = \frac{1}{n}\sum_{i:y_i=c} (\sigma( \vw^\top\rvx_i + b)-\ry_i)\left[\begin{array}{c}\rvx_i\\ 1\end{array}\right],
\quad \mu_x = \frac{1}{n}\sum_{i:y_i=c} \rvx_i.
\]
Due to the underrepresentation of the minority group in the data, the feature space center will be heavily biased towards the majority group which could hinder the clustering as illustrated in Figure \ref{fig:cosine}(a). 
On the other hand, the expression of $\mu_f$ above implies that gradient space center upweighs high-loss points which are more representative of the minority groups, resulting in a center in-between minority and majority groups. 
Thus, centering in the gradient space facilitates learning group annotations via clustering with the cosine distance as illustrated in Figure \ref{fig:cosine}(b). 
\end{itemize}

\begin{algorithm2e}[t]
    \SetKwInOut{Input}{Input}
    \SetKwInOut{Output}{Output}
\Input{DBSCAN hyperparameters $\eps$ and $m$}
Train the ERM classifier $\btt_0 \leftarrow \arg\min_{\btt \in \Theta'} \sum_{\rvz \in \gD} \ell (\ry, h_{\btt}(\vx))$ \;
\For{$\rvz \in \gD$}{
Compute its gradient $\vf \leftarrow \frac{\partial \ell (\ry, h_{\btt}(\rvx))}{\partial \btt} \mid_{\btt = \btt_0}$\;
}{}
% Collect the gradients $\set{\rvg_i}_{i=1}^{n_{\text{train}}}$ and $\set{\rvg_j}_{j=1}^{n_{\text{valid}}}$ of the samples from $\gD$ and $\val$, respectively\;
\For{$y \in \gY$}{
Consider all samples $\set{\rvz_i} \subset \gD$ with $\ervy = y$ and their corresponding gradients $\set{\vf_i}$\;
$\mu_f \leftarrow \operatorname{mean}(\set{\vf_i})$, compute the distance matrix $D$, where $D_{ij} = 1 - \frac{\langle \vf_i -\mu_f, \vf_j -\mu_f \rangle}{\norm{\vf_i-\mu_f}\cdot\norm{\vf_j-\mu_f}}$\;
% Compute the distance matrix $D$, where  $D_{ij} = \norm{\vf_i - \vf_j}_2$\;
Assign group annotations and identify outliers by performing DBSCAN clustering in gradient space: $\set{\hat{\ervg}_i} \leftarrow \operatorname{DBSCAN}(D,\eps,m)$, where $\hat{\ervg}_i = -1$ indicates outliers\;
}{}
% $\gD' \leftarrow \set{(\rvx, \ervg, \ervy)}_{\set{\ervg \neq -1, \rvz \in \gD}}$.
% Perform DRO on $\train'$ with validation set $\val'$\;
\Output{Dataset with predicted group annotations $\gD' \leftarrow \set{(\rvx, \hat{\rg}, \ry)}_{\set{\hat{\rg} \neq -1, \rvz \in \gD}}$, where the detected outliers are removed}
  \caption{\grasp{}}
 \label{alg:grass}
\end{algorithm2e}

\textbf{Quantitative comparison.}
% Motivated by the toy example above, we further showcase the superiority of gradient representations by comparing clustering results in feature space and gradient space. 
We compare the group identification quality of clustering in feature space and gradient space on two datasets consisting of 4 groups. We consider both clean and contaminated versions. The first dataset is Synthetic based on the Figure \ref{fig:motivating} illustration. The second dataset is known as Waterbirds \citep{sagawa2019distributionally}. It is a semi-synthetic dataset of images of two types of birds placed on two types of backgrounds. We embed the images with a pre-trained ResNet50 \citep{he2016resnet} model. 
% Both datasets have 4 groups. 
% Experiments are performed on a synthetic dataset and the Waterbirds dataset, each with a clean version and a contaminated version.  
To obtain gradient space representations, we trained logistic regression models. See Section~\ref{sec:exp} for additional details. 

We consider three popular clustering methods: K-means, DBSCAN with Euclidean distance, and DBSCAN with centered cosine distance.
Group annotations quality is evaluated using the Adjusted Rand Index (ARI) \citep{hubert1985ari}, a measure of clustering quality. 
Higher ARI indicates higher group annotations quality, and ARI~$=1$ implies the predicted group partition is identical to the true group partition. 
The definition of ARI is provided in Appendix~\ref{app:background}. We summarize the results in Table~\ref{tab:representation}.
% \kl{Give the definition in the appendix.} 
% We measure the group inference performance with Adjusted Rand Index (ARI)\yuchen{cite}, which is a popular metric for evaluating the clustering results. 
 % presents the performance of group inference by feature space clustering and gradient space clustering.
Clustering in the gradient space noticeably outperforms clustering in the feature space.
These results provide empirical evidence that gradient space facilitates learning of group annotations via clustering.
Visualization of the feature space and gradient space of the Synthetic and Waterbirds datasets are provided in Appendix~\ref{sec:space_vis}.

% \begin{wrapfigure}{r}{0.5\textwidth}
%     \centering
%     \includegraphics[width=0.5\textwidth]{iclr2023/figures/grad_vis_waterbirds.pdf}
%     \caption{Visualization of the gradient space. \yuchen{waterbirds dataset, more details to be added...}}
%     \label{fig:grad_vis}
% \end{wrapfigure}

% \begin{wraptable}{r}{0.55\textwidth}
\begin{table}[t]
\caption{
\textbf{Group identification quality of clustering methods in feature space and gradient space measured by Adjusted Rand Index (ARI). }
Higher ARI indicated higher group identification quality. 
The results are reported on clean and contaminated versions of Synthetic and Waterbirds datasets. 
Three different clustering methods are considered: K-means, DBSCAN \wrt{} Euclidean distance (DBSCAN/Euclidean), and DBSCAN \wrt{} centered cosine distance (DBSCAN/Cos). 
We set $k=2$ for K-means, which is the number of groups per class in these datasets. 
The gradient space clustering noticeably outperforms feature space clustering.}
% \kl{what if you extract features using some  kind of feature extractor? (instead of raw input features)}

\label{tab:representation}
\begin{center}
\begin{small}
\begin{sc}
\tiny{
\begin{tabularx}{\textwidth}{lc|ccc|ccc}
% \begin{tabularx}{ccccc} % \diagbox{Dataset}{Method}
\toprule
  \multirow{2}{*}{Dataset} & \multirow{2}{*}{Outliers?} & \multicolumn{3}{c}{Feature Space} & \multicolumn{3}{c}{Gradient Space} \\ 
 & & K-means & DBSCAN/Euclidean & DBSCAN/Cos & K-means & DBSCAN/Euclidean & DBSCAN/Cos \\ \midrule
 \multirow{2}{*}{Synthetic} & \xmark & .5505 & .5923 & .5133 & \textbf{.8409} & .7724 & .6943  \\
 & \cmark & .3631 & .6042 & .4946 & .6436 & \textbf{.7237} & .6944\\ \midrule
 \multirow{2}{*}{Waterbirds} & \xmark & .3932 & .0000 & .0418 & .7235 & .7304 & \textbf{.7453} \\
 & \cmark & .3932 & .0000 & .0418 & .7171 & .7304 & \textbf{.7453} \\
\bottomrule
\end{tabularx} }
\end{sc}
\end{small}
\end{center}
 % \vspace{-0.3in}
% \end{wraptable}
\end{table}

\subsection{\grasp{} for Group Inference and Outlier Identification}\label{sec:grasp_alg}
% Due to the similarity of tail distribution and outliers, the main challenge is to identify them simultaneously. 
% We provide an example to illustrate the similarity here.
% One common method for outlier detection is classifying samples with higher loss as outliers. 
% However, the samples from the clean tail distribution also suffer from higher loss, which makes it difficult to distinguish them from outliers.
% Therefore, instead of utilizing loss or input features, we leverage the gradient, which contains higher-dimensional information of the data for identifying tail samples and outliers. 
% Intuitively, the gradients within the same subgroup behave similarly while outlier gradients do not due to the lack of structure among true outliers. 
%Given our findings in Sec.~\ref{sec:space} that gradient space makes it easier for clustering and outlier identification, here we present our Gradient Space Partition (\grasp{}) in more details. 
Having motivated our choice of performing clustering in the gradient space, we now present \grasp{} in detail.
We then describe how to train a distributionally and outlier robust model using \grasp{}.

\textbf{Clustering method and distnace measure.} Results in Table \ref{tab:representation} indicate that both K-means and DBSCAN perform well in the gradient space. DBSCAN is a density-based clustering algorithm, where clusters are defined as areas of higher density, while the rest of the data is considered outliers. In this work, we choose to use DBSCAN for its ability to identify outliers, which is an important aspect of the problem we consider. As an additional benefit, unlike K-means, it does not require knowledge of the number of groups. See Appendix \ref{app:background} for a detailed description of DBSCAN.

In terms of distance measure, we recommend cosine distance due to its better performance on the Waterbirds data, which closer resembles real data. We note that the distance and clustering method choices could be reconsidered depending on the application. For example, for Gaussian-like data without outliers, K-means performed better in Table \ref{tab:representation}.

% While one can also use other clustering methods such as K-means
% \kl{K-means -- please find and replace it everywhere}
% , we suggest using DBSCAN~\citep{ester1996dbscan} as the method to perform clustering in gradient space. 
% DBSCAN is a density-based clustering algorithm, where clusters are defined as areas of higher density than the reminder of the dataset. 
% Therefore, it can learn clusters of different shapes and sizes without knowing the number of clusters and detect outliers, while K-means cannot.
% Besides, the results reported in Table.~\ref{tab:representation} also highlight the benefits of DBSCAN in terms of clustering the gradients and performing group inference.
% We provide more details of DBSCAN in Sec.~\ref{app:background}.

\textbf{\grasp{}.} We present the pseudocode of \grasp{} in Algorithm~\ref{alg:grass}.
We first train an ERM classifier $h_{\btt}(\cdot)$ and collect the gradients of sample's loss \wrt~model parameters $\btt$. 
We then compute the pairwise centered cosine distances within each class $y\in\gY$ using gradient representations, as discussed in Sec.~\ref{sec:space}.
% When using cosine similarity, we centralize the gradients before computing the cosine similarity distance matrix, as discussed in Sec.~\ref{sec:space}. 
Lastly, to estimate the group annotations and identify outliers, we apply DBSCAN on these distance matrices for each class $y\in \gY$.
% giving the group annotations and detected outliers. 

\textbf{Training models with improved worst-group performance in the presence of outliers using \grasp{}.}
% Next, we remove the outliers and use the predicted spurious subgroup information for solving \eqref{opt:dro}.
% Note that the goal of gradient clustering is not just to predict the subgroup information correctly, but also to identify outliers. 
We discard the identified outliers and then provide learned group annotations as inputs to a Group-aware method of choice. For concreteness, in this work, we will use \eqref{opt:gdro}. Specifically, we employ 
 the method of \citet{sagawa2019distributionally} to solve the gDRO problem, which is a stochastic optimization algorithm with convergence guarantees. We note that other choices could be appropriate. For example, methods accounting for noise in group annotations~\citep{lamy2019noise,mozannar2020fair,celis2021fair}
 %\my{Yuchen, please cite relevant ones} 
 are interesting to consider as they could counteract mistakes in \grasp{} annotations.
 
\paragraph{Remark.} We note that 
the model $h_\theta$ and parameter space 
$\Theta$ used for computing gradient representations $\vf$ and learning group annotations with \grasp{} can be different from the classifier and parameter space used for the final model training. 
For example, one can train a logistic regression model (using features from a pre-trained model when appropriate) and collect the corresponding gradients for \grasp{}, and then train a deep neural network of choice with the estimated group annotations.
% as the final model. % \my{last part is not related to this section}

\section{Experiments}\label{sec:exp}
In this section, we conduct extensive experiments on both synthetic and benchmark datasets to evaluate the performance of \grasp{}.% \footnote{Our code is available in anonymous Github repository~\url{https://anonymous.4open.science/r/private_demographics-DB5A/}.}
\footnote{Our code is available in Github repository~\url{https://github.com/yzeng58/private_demographics}.}
Our results show that \grasp{} outperforms the state-of-the-art baselines in terms of group identification quality and downstream worst-group performance while providing robustness to outliers.

\subsection{Datasets and baselines}
%In our experiments, we study four datasets, including both synthetic and benchmark datasets.

\textbf{Synthetic.} We generate a synthetic dataset of 1,000 samples with two features $\rvx\in\sR^2$, a group attribute $\rg\in[4]$, and a binary label $\ry \in \set{0,1}$, similar to the motivating example of Figure~\ref{fig:motivating}.  
\textbf{(Clean):} The synthetic dataset consists of 10 Gaussian clusters with a variance of 0.01, and each Gaussian cluster contains 100 samples.
Class 0 is divided into two groups: group 3 consists of four Gaussian clusters with centers $(1,5),(1,3),(1,2),(1,1)$; group 2 consists of one Gaussian cluster with center $(0,4)$.
Similarly, Class 1 is divided into two groups: group 1 consists of four Gaussian clusters with centers $(0,5),(0,3),(0,2),(0,1)$; group 2 consists of one Gaussian cluster with center $(1,4)$. 
\textbf{(Contaminated):} We contaminate the synthetic dataset by flipping randomly selected 5\% of labels.
The contaminated synthetic dataset is visualized in Appendix Figure~\ref{fig:syn_outlier_vis}. 
% adding $20$ outliers into the training dataset.
% These outliers are generated as $\rvx\sim \gN((.5,2.5)^\top, 10\mI_2), \ry\sim \operatorname{Bern}(.5)$, where $\mI_2\in \sR^{2\times 2}$ is the identity matrix. The resulting dataset is visualized in Appendix Figure \my{XXX: Yuchen please add a figure}

% We contaminate the synthetic dataset by adding $20$ outliers with group annotations $\rg = 5,6$ into the training dataset.
% These outliers $\rvz=(\rvx,\ry)$ are generated as $\rvx\sim \gN((.5,2.5)^\top, 10\mI_2), \ry\sim \operatorname{Bern}(.5), \rg = \ry+5$, where $\mI_2\in \sR^{2\times 2}$ is the identity matrix. 

\textbf{Waterbirds.} \textbf{(Clean):} Waterbirds \citep{sagawa2019distributionally,wah2011cub} is a semi-synthetic image dataset of land birds and water birds \citep{wah2011cub} placed on either land or water backgrounds using images from the Places dataset \citep{zhou2017places}. There are 11,788 images of birds on their typical (majority) and atypical (minority) backgrounds.
The task is to predict the types of birds and the background type is the group (2 background types per class, a total of 4 groups).
We follow an identical procedure to \citet{idrissi2022balancing} to pre-process the dataset.
\textbf{(Contaminated):} We contaminate the Waterbirds dataset by introducing outliers in the training and validation datasets. 
We flip the class labels of 2\% of the data, transform 1\% of the images with Gaussian blurring, color dither (randomly change the brightness, saturation, and contrast of the images) 1\% of the images, and posterize 1\% of the images maintaining 4 bits per color channel. 
We visualize a contaminated example in Appendix Figure~\ref{fig:waterbirds_process}.

% \kl{We could have run more experiments, i.e., changing the ratio of outliers}

\textbf{COMPAS \& CivilComments.} Both datasets are real and collected by humans, therefore likely to contain outliers.
\textbf{(COMPAS):} COMPAS \citep{compas} is a recidivism risk score prediction dataset consisting of 7,214 samples.
Each class $y\in[0,1]$ is divided into six groups: Caucasian males, Caucasian females, African-American males, African-American females, males of other races, and females of other races, making 12 groups in total. 
% These true group labelings are not available to \grasp{} and are used only for performance evaluation.
%This implies we divide the dataset into 12 groups. 
\textbf{(CivilComments):} CivilComments \citep{dixon2018measuring,koh2021wilds} is a language dataset containing online forum comments. 
The task is to predict whether comments are toxic or not. % {\color{red}KG: Comments on groups here?}
We follow a procedure identical to \citet{idrissi2022balancing} to preprocess the dataset.
We divide comments in each class into two groups according to the presence or absence of identity terms pertaining to protected groups (LGBT, Black, White, Christian, Muslim, other religion).

% \subsection{Baselines}
\textbf{Experimental baselines.} We compare \grasp{} to four different types of baselines: (1) standard empirical risk minimization (ERM), (2) a group-aware method (gDRO~\citep{sagawa2019distributionally}), (3) a group-oblivious method (DORO, CVaR-DORO variation~\citep{zhai2021doro}), and (4) two group-learning methods (EIIL~\citep{creager2021eiil}, George~\citep{sohoni2020george}). We chose DORO among the methods relying on loss values to improve worst-group performance because it is the only method from this group designed to be robust to outliers. Recall that only the group-aware method (gDRO) has access to the true group annotations, thus it should be interpreted as an ``oracle'' baseline.

We also perform an ablation study by considering an additional group-learning baseline, Feature Space Partitioning (FeaSP). It is identical to GraSP except it performs DBSCAN clustering in the feature space. Comparison to FeaSP emphasizes the importance of clustering in the gradient space as opposed to other choices such as the clustering method and distance measure.

\begin{table}[t]
\caption{
\textbf{Group identification performance of group-learning methods measured by Adjusted Rand Index (ARI). }
Higher ARI indicated higher group identification quality. 
The results are reported on clean and contaminated versions of Synthetic and Waterbirds datasets, COMPAS and CivilComments datasets. 
\grasp{} significantly outperforms the other group-learning baselines on all the tested datasets.
Moreover, we observe that \grasp{} is robust to outliers. 
% \yuchen{FeaSP on CivilComments has not been finished yet.}
}
\label{tab:ari}
% \vskip 0.15in
\begin{center}
\begin{small}
%\tiny{
\begin{sc}

\begin{tabularx}{.9\textwidth}{lcccccc}
\toprule
& \multicolumn{2}{c}{Synthetic} & \multicolumn{2}{c}{Waterbirds} & \multirow{2}{*}{COMPAS} &  \multirow{2}{*}{CivilComments}\\
Outliers? & \xmark & \cmark & \xmark & \cmark & &  \\
Method & ARI & ARI & ARI& ARI& ARI& ARI \\
\midrule
EIIL & -.0069& -.0043 & .0114 & .0078 & -.0025 & -.0001  \\
George&.6027& .4565 & .2832 & .2600 & .1962 & .1422  \\
FeaSP & .5133 & .4946 & .0418 & .0418 & .2956 & .2093 \\ 
\grasp{} (Ours) & \textbf{.6943} & \textbf{.6944} & \textbf{.7453} & \textbf{.7453} & \textbf{.5453} & \textbf{.2639}\\
\bottomrule
\end{tabularx}
\end{sc}
% }
\end{small}
\end{center}
\vskip -0.15in
\end{table}

\subsection{Evaluation of \grasp{}}
In this section, we assess the performance of \grasp{} in terms of group identification and downstream tasks of training models with comparable performance across groups, both with and without outliers.
% Although most of existing work~\citep{creager2021eiil,zhai2021doro} assumes a validation dataset with known group annotations for hyperparameters selection,
In all experiments, we consider true group annotations unknown in both train and validation data (except for ``oracle'' gDRO which has access to true group annotations in both train and validation data). Arguably, this setting is more practical due to group annotations often being expensive or infeasible to obtain, even for a smaller validation set. We note that this setting differs from the majority of prior works considering unknown group annotations (see Table \ref{tab:baseline}). For example, inspecting Table~5 in Appendix~B.2 of \citet{zhai2021doro}, we notice that their DORO is unable to improve upon ERM without access to validation data with true group annotations (see results for non-oracle model selection). We report results with known validation group annotations in Appendix~\ref{sec:w_val}.

% To better simulate a real-world scenario, all the experiments presented in the main text are conducted without known validation group annotations (except ``oracle'' gDRO which has access to true group annotations in both train and validation data); this makes the task more challenging. 
% We defer the experiment results reported on known validation group annotations to Sec.~\ref{sec:w_val} in the supplement.
% We study how hyperparameters affect the performance of \grasp{} and provide details of hyperparameter selection for the two experiments in Section~\ref{sec:hyperparameter}. 
%, and follow the preprocessing steps of \citet{idrissi2022balancing} for the Waterbirds and CivilComments datasets.

\textbf{Group annotations quality.} 
The first experiment examines the quality of group annotations learned with \grasp{}. 
To collect the gradients of the data's losses \wrt{} the model parameters, we train a logistic regression model on the Synthetic dataset, a three-layer ReLU neural network with 50 hidden neurons per layer on the COMPAS dataset, and a BERT~\citep{devlin2018bert} model on the CivilComments dataset (due to the large number of parameters in BERT, we only consider the last transformer and the subsequent prediction layer when extracting gradients). 
For the Waterbirds dataset, we first featurize the images using a ResNet50 pre-trained on ImageNet \citep{deng2009imagenet}, and then train a logistic regression. 
We then use DBSCAN clustering with centered cosine distance. We select DBSCAN hyperparameters using standard clustering metrics that do not require knowledge of the true group annotations, see Appendix~\ref{sec:hyperparameter}.
% , where we demonstrate the \grasp{} (in is fairly robust to DBSCAN hyperparameters.
% We compare to three group learning baselines: EIIL, George, and FeaSP.

% computed on the centralized and normalized gradients. 

In Table~\ref{tab:ari}, we compare group identification quality of \grasp{} (measured with ARI) to three group learning baselines, EIIL, George, and FeaSP, across four datasets. There are two key observations supporting the claims made in this paper: (i) clustering in the gradient space (\grasp{}) outperforms clustering in the feature space (FeaSP and George), as well as other baselines (EIIL); (ii) \grasp{} is robust to outliers, i.e. it performs equally well in the presence and absence of outliers. To comment on the low ARI of EIIL, we note that the Invariant Risk criteria EIIL optimizes was designed primarily for invariant learning (i.e., learning environment labels) \citep{arjovsky2019invariant,creager2021eiil}, which may not be suitable for learning group annotations.
% ARI for all methods and datasets\yuchen{, except FeaSP on CivilComments dataset -- how should I say this to make it less weird?}.

% First, we observe the low performance of EIIL in group inference. 
% This is as expected, as EIIL is not explicitly designed for identifying true groups but groups that maximally informative for downstream tasks.
% Meanwhile, the additional outliers added to the Synthetic and Waterbirds datasets degrade the ARI of George and FeaSP, while the performance of \grasp{} remains largely unchanged.
% More importantly, \grasp{} significantly outperforms the other three baselines in \emph{all} datasets for this group inference task. 
% Given that \grasp{} does not require validation group annotations, this result implies that the worst-group performance estimated by \grasp{} can be a good proxy for model selection. 
% \yuchen{More experiments? Open direction?}

\textbf{Worst-group performance.} This is the standard metric when comparing methods for training ML models with comparable performance across groups (evaluated \wrt true group annotations) \citep{sagawa2019distributionally,koh2021wilds}. For the group-learning methods (\grasp{}, FeaSP, George, EIIL), we first discard identified outliers if applicable (\grasp{} and FeaSP), and then train gDRO with the corresponding learned group annotations. We also use the learned group annotations on the validation data to select the corresponding gDRO hyperparameters. In Appendix~\ref{sec:robust_cluster} we demonstrate that \grasp{} worst-group performance is fairly robust to the corresponding DBSCAN hyperparameters. For ERM and DORO we used the validation set overall performance for hyperparameter selection.

% For ERM and DORO we used the validation set overall performance for hyperparameter selection. We highlight the utility of 

% ERM and DORO; good to learn groups and do it well.

For all methods, on a given dataset, we train models with the same architecture and initialization. Recall that these models can be different from the models used in estimating group annotations with any of the group-learning methods. See Appendix~\ref{sec:hyperparameter} for details.

We summarize results in Table~\ref{tab:worst}. \grasp{} outperforms baselines on Synthetic, COMPAS, and CivilComments datasets. For the Waterbirds dataset, \grasp{} also performs relatively well. Interestingly, EIIL performs best on the contaminated Waterbirds dataset, despite the poor ARI discussed earlier. It is, however, failing on the COMPAS dataset. We also notice that \grasp{} outperforms ``oracle'' gDRO on Synthetic and COMPAS datasets. This could be due to the fact that gradient space clustering helps to focus on ``harder'' instances, as discussed in Section~\ref{sec:space}, while the available (``oracle'') group annotations (at least on COMPAS), might be noisy.

\begin{table}[t]
\caption{
\textbf{Downstream worst-group accuracy and average accuracy on the test data.}
% {\color{red} KG: what do these arrows mean?} \yuchen{The values of the two metrics are the  higher the better :) }}
The average test accuracy is a re-weighted average of the group-specific accuracies, where the weights are based on the training distribution.
The results are reported on clean and contaminated versions of Synthetic and Waterbirds datasets, COMPAS and CivilComments datasets. 
% The gDRO (oracle) here is a Group-aware method, which can get access to the true group annotations.
We observe that \grasp{} significantly outperforms the group-oblivious (DORO) and other group-learning approaches (EIIL, George, FeaSP) methods on Synthetic, COMPAS, and CivilComments datasets, and performs relatively well on Waterbirds datasets, while being robust to outliers. 
% Note that \grasp{} sometimes outperforms gDRO (oracle), which can get access to the true group annotations.
% This is because \grasp{} may focus on ``harder'' instances more, which potentially affect the results most.
% \yuchen{How to write the excuse for this missing entry?}{\color{red} KG: I don't really think there can be an excuse that reviewers would allow here (they will say the paper is not ready). Probably best to simply leave blank without explanation, and provide when the reviewers ask}
}
\label{tab:worst}
% \vskip 0.15in
\begin{center}
\begin{small}
\tiny{
\begin{sc}
\setlength{\tabcolsep}{8pt}
\begin{tabularx}{\textwidth}{lcccccc}
\toprule
& \multicolumn{2}{c}{Synthetic} & \multicolumn{2}{c}{Waterbirds} & \multirow{2}{*}{COMPAS} &  \multirow{2}{*}{CivilComments}\\
Outliers? & \xmark & \cmark & \xmark & \cmark & &  \\
Method & Worst.(Avg.) & Worst.(Avg.) & Worst.(Avg.)& Worst.(Avg.)& Worst.(Avg.)& Worst.(Avg.) \\
\midrule
ERM & .6667(.8823) & .5333(.8273) & .6075(.9673) & .5249(.9621) & .4706(.6792) & .4659(.9213) \\ \midrule
DORO & .6667(.8823) & .6000(.8342) & .5888(.9694) & .6636(.9686) & .4706(.6801) & .4905(.9182)\\ 
EIIL & .6667(.8783) & .6000 (.8115) & .6916 (.9645) & \textbf{.7056(.9629)} & .0588 (.6046) & .6056(.9066) \\
George &.5333(.8732) &  .6000(.8342) & .\textbf{7523}(.9612) & .5897(.9100) & .4416(.6232) & .5897(.9100)\\
FeaSP & .6667(.8823) & .6667(.8823)& .1417(.9346) & .1417(.9346) & .4416(.6232) & .6056(.9066) \\ 
\grasp{} (Ours) & \textbf{.8000}(.8926)& \textbf{.8000}(.8926) & .6854(.9654) & .6798(.9004) & \textbf{.4743}(.6717) & \textbf{.6798(.9004)} \\ \midrule
gDRO (oracle) & .7333(.8639) & .8000(.8755) & .8665(.9272) & .8545(.9081) & .4625(.6807) & .6941(.8767) \\
\bottomrule
\end{tabularx}
\end{sc}
}
\end{small}
\end{center}
\vskip -0.15in
\end{table}

% \section{Additional Related Work}
% {\color{red} KG: Maybe this is already redundant with above intro/background discussions, if not maybe can be merged into the above.}
% \paragraph{Subgroup Shift} Subgroup shift refers to the discrepancy between the proportion of subgroups in the training distribution and the test distribution.
% A model robust to subgroup shift basically means this model performs well on all subgroups....

% \paragraph{Outlier-robust DRO} ...

% \paragraph{Gradient-based representations}
%\section{Discussion}

\section{Conclusion}
%{\color{green} KG: I quickly drafted the below, it may not be fully accurate}

In this work, we considered the problem of learning group annotations in the presence of outliers. Our method allows training models with comparable performance across groups to alleviate spurious correlations and accommodate subpopulation shifts when group annotations are not available and need to be estimated from data. We accomplished this by leveraging existing outlier-robust clustering approaches to estimate the group (and outlier) memberships of each point in the dataset. Key to our proposed approach is performing the clustering in the \emph{gradient space}, where the gradient is of the loss at each point with respect to model parameters. We provided strong intuitive and empirical justifications for using the gradient space over the feature space.
Finally, we provided a variety of synthetic and real-world experiments where \grasp{} consistently outperformed or nearly matched the performance of all comparable baselines in terms of both learned group annotations quality and downstream worst-group performance.
% In particular, \grasp{} achieves robustness to outliers where all preexisting baselines fail.
% \justin{open problems? next steps?}

One advantage of the gradient space is the simplification of the structure of the correctly classified points (often the majority group), which is also a limitation if identifying \emph{subgroups} within the majority group is of interest. This does not affect the downstream worst-group performance, but may be undesirable from the exploratory data analysis perspective.

As a next step, when training models with \grasp{} group annotations, it would be interesting to consider alternatives to gDRO that are accounting for noise in group annotations \citep{lamy2019noise,mozannar2020fair,celis2021fair} to counteract \grasp{} estimation error. Alternatively, one can consider training with group-oblivious methods such as DORO \citep{zhai2021doro} and performing model selection on the validation data with \grasp{} group annotations.

\section*{Acknowledgments}
% We thank Luann Jung for preliminary work on \grasp{}.
% This work is based on of Luann Jung's Master thesis, which conducts preliminary experiments on \grasp{}.
We thank Luann Jung for conducting a preliminary study of \grasp{} for her Master's thesis.
Kangwook Lee was supported by NSF/Intel Partnership on Machine Learning for Wireless Networking Program under Grant No. CNS-2003129 and by the Understanding and Reducing Inequalities Initiative of the University of Wisconsin-Madison, Office of the Vice Chancellor for Research and Graduate Education with funding from the Wisconsin Alumni Research Foundation.
The MIT Geometric Data Processing group acknowledges the generous support of Army Research Office grants W911NF2010168 and W911NF2110293, of Air Force Office of Scientific Research award FA9550-19-1-031, of National Science Foundation grants IIS-1838071 and CHS-1955697, from the CSAIL Systems that Learn program, from the MIT--IBM Watson AI Laboratory, from the Toyota--CSAIL Joint Research Center, and from a gift from Adobe Systems.
\bibliography{refs}
\bibliographystyle{iclr2023_conference}
\clearpage
\appendix

\vspace{-.4in}
\section{Background}\label{app:background}
In this section, we provide the details of Adjusted Rand Index (ARI), and describe the complete algorithm of DBSCAN.

\paragraph{Adjusted Rand Index (ARI)}~\citep{hubert1985ari}
The Adjusted Rand Index (ARI) is a measure of the degree of agreement between two data partitions and accounts for the chance grouping of elements in the data sets.
In our case, consider true group partition $P$ and estimated group partition $\hat{P}$.
ARI can be computed by
\begin{equation}
	ARI(P,\hat{P}) = \frac{\sum_{k, k'}\left(\begin{array}{c}
			n_{k k'} \\
			2
		\end{array}\right)-\left[\sum_k\left(\begin{array}{c}
			n_k \\
			2
		\end{array}\right) \sum_k'\left(\begin{array}{c}
			n_{k'} \\
			2
		\end{array}\right)\right] /\left(\begin{array}{l}
			n \\
			2
		\end{array}\right)}{\frac{1}{2}\left[\sum_k\left(\begin{array}{c}
			n_k \\
			2
		\end{array}\right)+\sum_k'\left(\begin{array}{c}
			n_{k'} \\
			2
		\end{array}\right)\right]-\left[\sum_k\left(\begin{array}{c}
			n_k \\
			2
		\end{array}\right) \sum_k'\left(\begin{array}{c}
			n_k' \\
			2
		\end{array}\right)\right] /\left(\begin{array}{c}
			n \\
			2
		\end{array}\right)},
\end{equation}
where $n_{kk'}$ is the number of data points belonging to $\gG_k \in P$ assigned to group $\hat{\gG}_{k'} \in \hat{P}$, $n_k = |\gG_k|$, $n_{k'} = |\gG_{k'}|$, and $n$ is the total number of samples in the dataset.

\paragraph{DBSCAN}~\citep{ester1996dbscan}
% We begin by introducing the DBSCAN clustering algorithm~\citep{ester1996dbscan}---a clustering and outlier-detecting method that does not need the number of clusters to be known. 
% Assume that a distance matrix $D$ of the dataset is given.  
% For DBSCAN, a sample is called a ``core sample'' if there exist $m$ other samples within a distance of $\eps$ from this sample.
% DBSCAN starts from one cluster with one arbitrary core sample, and continues to add core samples from the neighborhood of the cluster to the cluster until all the core samples in the $\eps$-neighborhood of the cluster have been visited, and then add the rest of the samples in the $\eps$-neighborhood of the cluster into the cluster. 
% Next, DBSCAN creates another cluster with an arbitrary unvisited core samples and follows the same steps, introducing clusters and expanding them until all the core samples have been visited. 
% The remaining samples that are not added into any clusters are identified as outliers. 
% DBSCAN clustering requires two hyperparameters ($\eps, m$) and a distance matrix $D$ as input. \yuchen{Polish}
DBSCAN is a clustering and outlier-detecting method that does not require the number of clusters to be known. 
It operates on a distance matrix $D$.
We call a sample as a "core sample" if there exist $m$ other samples within a distance of $\eps$ from this sample. 
DBSCAN starts with a single cluster that contains an arbitrary core sample and adds core samples from the neighborhood of the cluster to the cluster until all core samples in the $\eps$-neighborhood of the cluster have been visited. 
It then adds the remaining samples in the $\eps$-neighborhood of the cluster to the cluster. 
Next, DBSCAN creates another cluster and expands that cluster by finding unvisited core samples. 
It then repeats this process of creating and expanding clusters until all core samples have been visited. 
Any remaining samples that are not added to a cluster are considered outliers.
Note that DBSCAN clustering requires two hyperparameters ($\eps, m$) and a distance matrix $D$ as input.

\section{Visualization of Gradient Space and Feature Space}\label{sec:space_vis}
\begin{figure}[h]
	\begin{subfigure}[b]{.49\textwidth}
		\includegraphics[width=\textwidth]{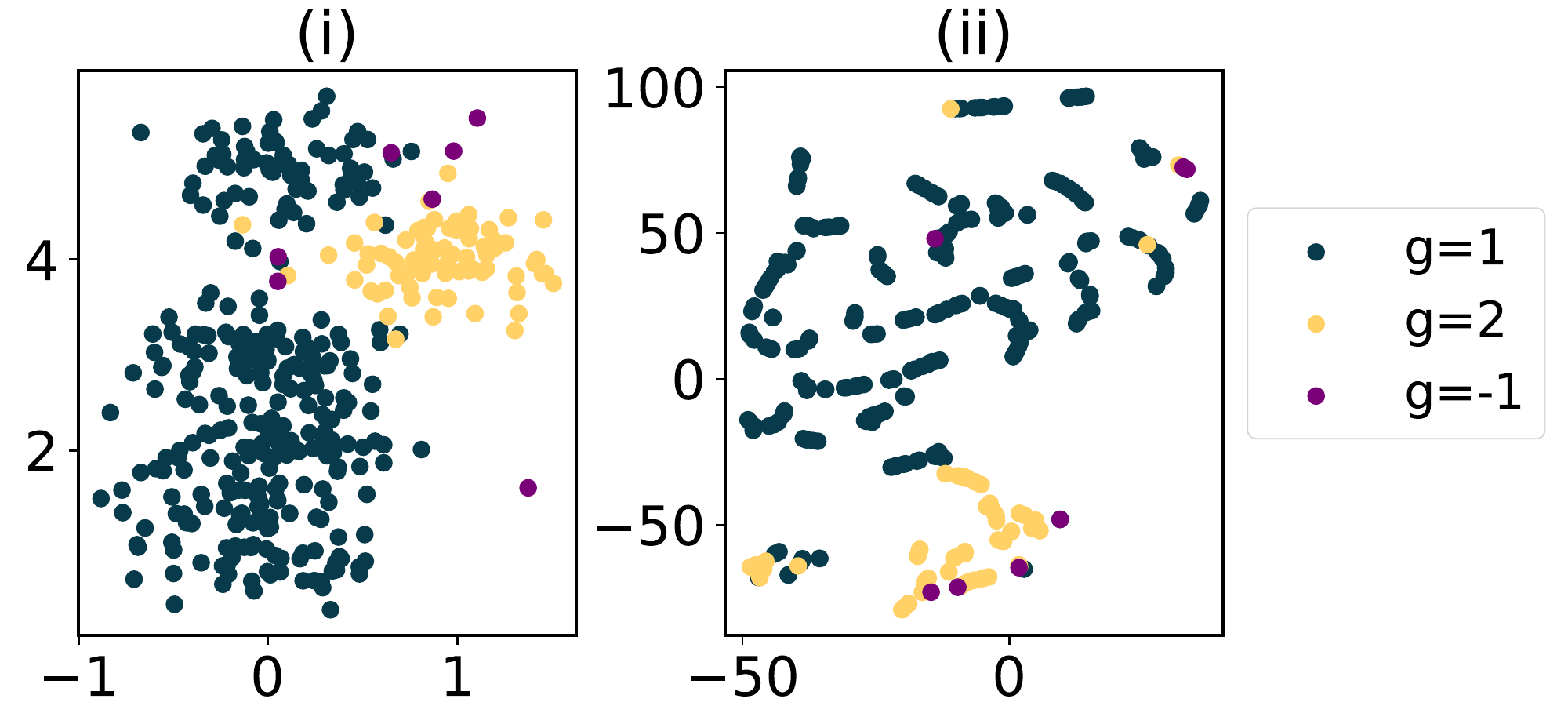}
		\caption{Visualization of input features and 2D t-SNE results of gradients with $y=0$ on contaminated Synthetic dataset. (i) Input features. (ii) 2D t-SNE results of gradients.}
		\label{fig:grad_syn}
	\end{subfigure} \hfill
	\begin{subfigure}[b]{.49\textwidth}
		\includegraphics[width=\textwidth]{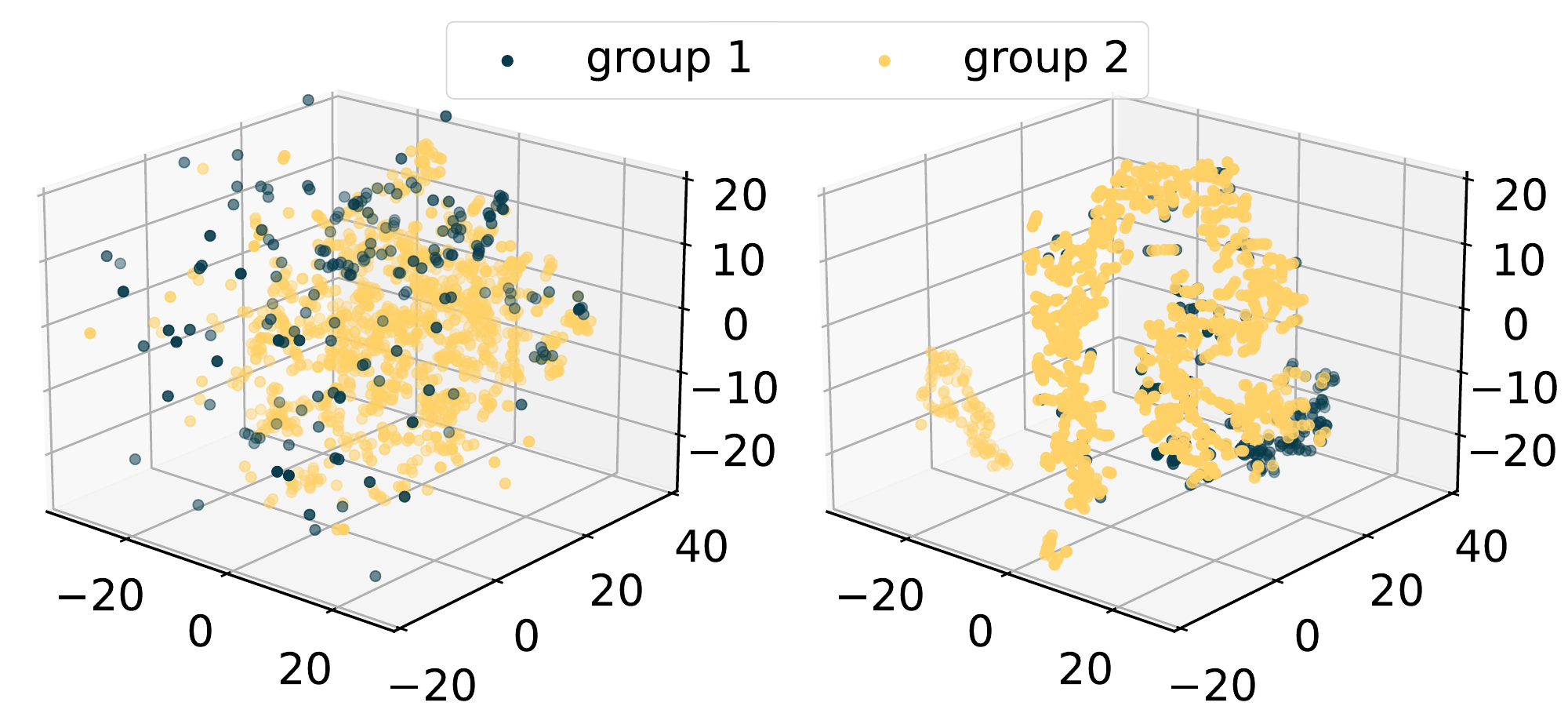}
		\caption{3D t-SNE visualization of features and gradients with $y=1$ on contaminated Waterbirds dataset. Left: 3D t-SNE visualization of features extracted from ResNet-50 pretrained on ImageNet~\citep{deng2009imagenet}. Right: 3D t-SNE visualization of gradients of the sample's loss \wrt{} the parameters of the last layer.}
		\label{fig:grad_waterbirds}
	\end{subfigure}
	% \caption{Visualization of input features and gradients of samples in contaminated Synthetic and Waterbirds datasets.}
\end{figure}

In this section, we visualize the gradient space and feature space of contaminated Synthetic (see Fig.~\ref{fig:grad_syn}) and Waterbirds dataset (see Fig.~\ref{fig:grad_waterbirds}).

\section{Experiment}

\begin{figure}[h]
	\begin{subfigure}[b]{.38\textwidth}
		\centering
		\includegraphics[width=\textwidth]{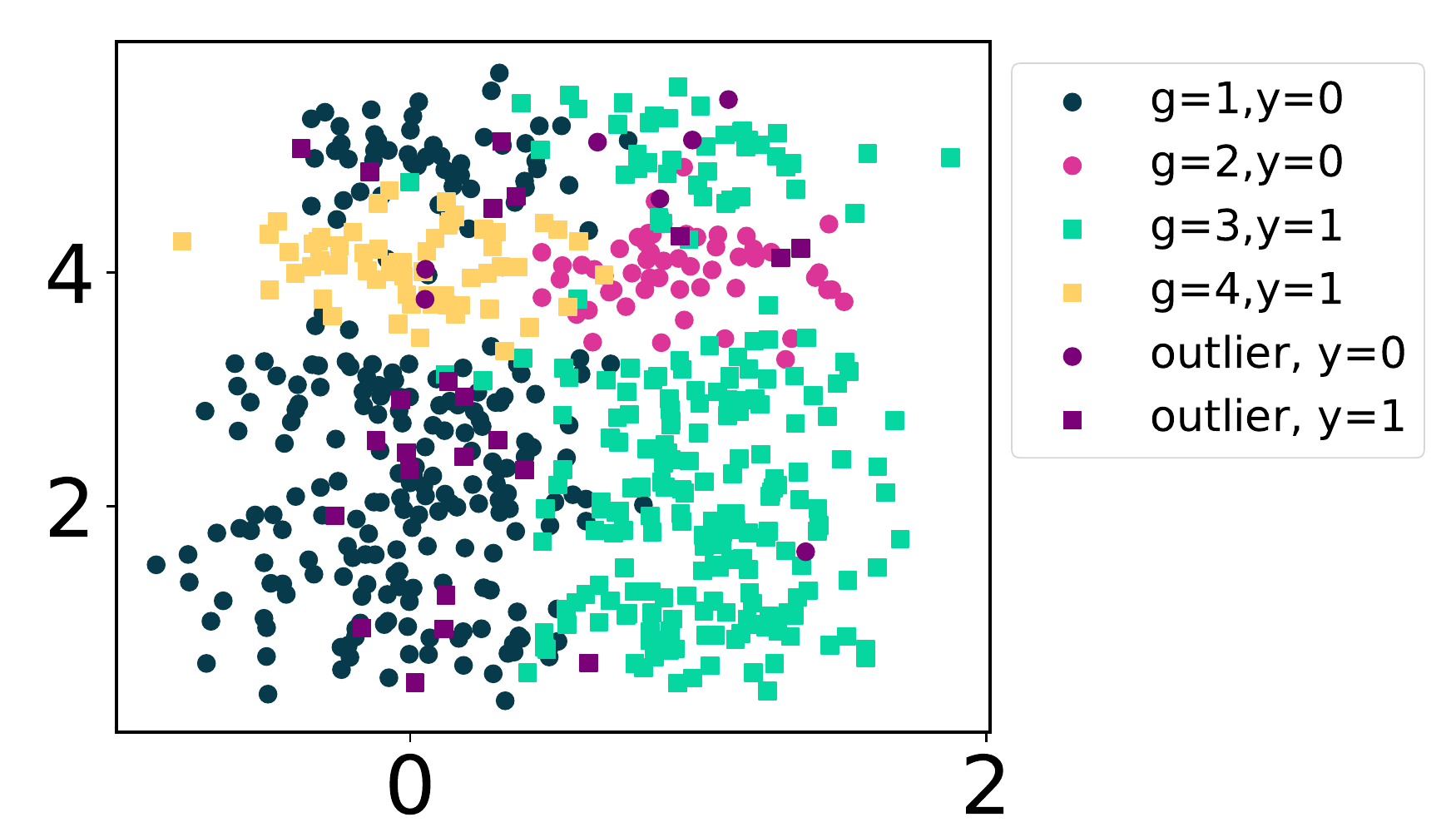}
		\caption{}
		\label{fig:syn_outlier_vis}
	\end{subfigure} \hfill
	\begin{subfigure}[b]{.3\textwidth}
		\includegraphics[width=\textwidth]{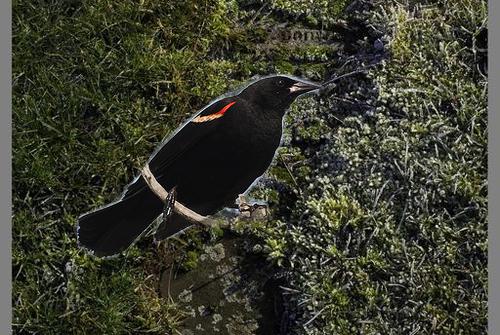}
		
		\caption{}
		\label{fig:waterbirds_original}
	\end{subfigure} \hfill
	\begin{subfigure}[b]{.3\textwidth}
		\includegraphics[width=\textwidth]{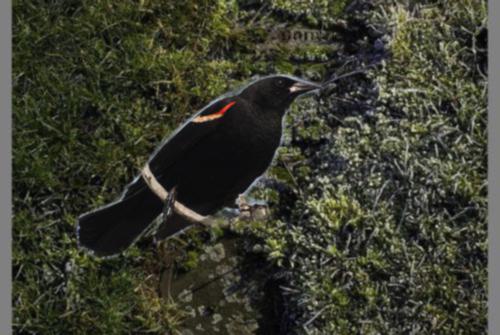}
		\caption{}
		\label{fig:waterbirds_process}
	\end{subfigure}
	\caption{(a) Scatter plot of contaminated Synthetic dataset. 
		(b) Original image of \texttt{010.Red\_winged\_Blackbird/Red\_Winged\_Blackbird\_0079\_4527.jpg}.
		(c) Image \texttt{010.Red\_winged\_Blackbird/Red\_Winged\_Blackbird\_0079\_4527.jpg} after Gaussian blurring.
	}
\end{figure}

\subsection{More Details of Experiment Setup}\label{sec:hyperparameter}

\begin{figure}[h]
	\centering
	\includegraphics[width=\textwidth]{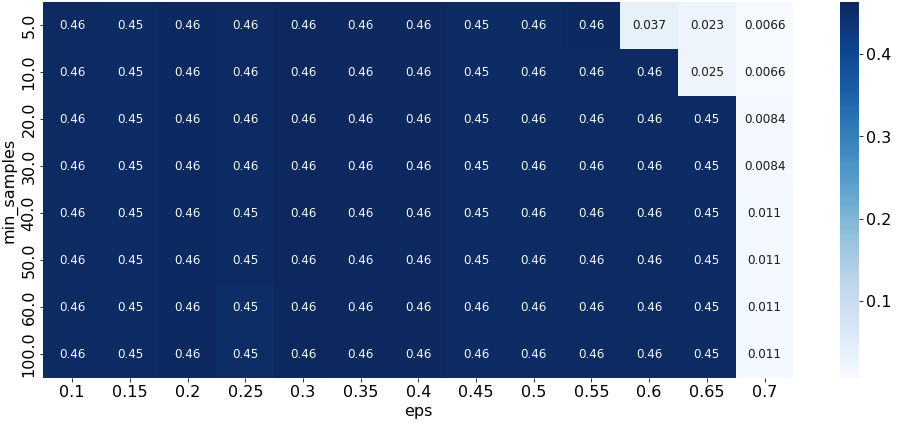}
	\caption{Group identification quality of \grasp{} v.s. DBSCAN clustering hyperparameters (\texttt{eps}: $\eps$, \texttt{min\_samples}: $m$) measured in Adjusted Rand Index (ARI) on class 0 of Waterbirds dataset.}
	\label{fig:waterbirds_heatmap_ari_0}
\end{figure}

\begin{figure}[h]
	\centering
	\includegraphics[width=\textwidth]{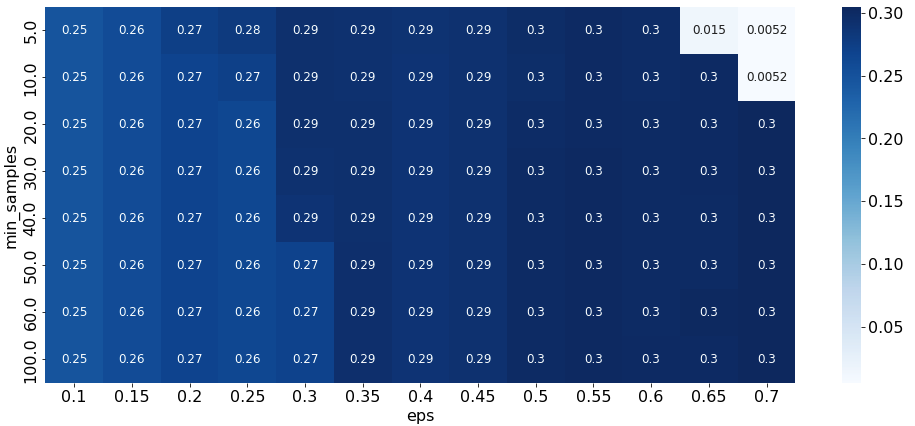}
	\caption{Group identification quality of \grasp{} v.s. DBSCAN clustering hyperparameters (\texttt{eps}: $\eps$, \texttt{min\_samples}: $m$) measured in Adjusted Rand Index (ARI) on class 1 of Waterbirds dataset.}
	\label{fig:waterbirds_heatmap_ari_1}
\end{figure}
\paragraph{Datasets}
The batch size of Synthetic, Waterbirds, COMPAS, and CivilComments datasets are 128, 128, 128, and 32 for both group inference and downstream DRO tasks. 
We split the Synthetic and COMPAS datasets into training, validation, and test datasets at the ratio of 0.6:0.2:0.2. 
We follow an identical procedure to \citet{idrissi2022balancing} to pre-process the Waterbirds and Civilcomments dataset.
Fig.~\ref{fig:syn_outlier_vis} visualizes the contaminated Synthetic dataset.
We provide an example of a contaminated sample in Fig.~\ref{fig:waterbirds_original} and Fig.~\ref{fig:waterbirds_process}, which present an image before and after Gaussian blurring. 

\paragraph{Group annotations quality.}
To collect the gradients of the corresponding datum's loss \wrt the model parameters, we train a logistic regression model on 50 epochs on the Synthetic dataset, a three-layer ReLU neural network with 50 hidden neurons for
300 epochs on the COMPAS dataset, and a BERT~\citep{devlin2018bert} model for 10 epochs on the
CivilComments dataset. For the Waterbirds dataset, we first featurize the images using a ResNet50
pre-trained on ImageNet~\citep{deng2009imagenet}, and then train a logistic regression for 360 epochs
We tune the DBSCAN clustering hyperparameters $\eps\in\set{.1,.2,.3,.5,.7}, m\in\set{10,20,30,50,70,100}$ for each $y\in \gY$, for both FeaSP and \grasp{}.
We tune the learning rate of EIIL in $\set{10^{-1}, 10^{-2}, 10^{-3}, 10^{-4}}$, run EIIL for 50 epochs on Synthetic, Waterbirds, and COMPAS datasets, three epochs on Civilcomments dataset. 
We tune the overcluster factor of George in $\set{1,2,5,10}$, and employ the over-cluster Gaussian Mixture Model clustering for George. 
Lastly, we select the best EIIL epoch and other hyperparameters based on \emph{Silhouette Coefficient}, a measure assessing the clustering quality in terms of the degree to which a sample clusters with other similar samples.
% \my{remaining sentences should go to appendix; also batch size of what? ERM models or final models?}

\paragraph{Worst-group performance.} 
We use Adam optimizer for all trainings. 
We tune outlier fraction $\eps \in \set{.005,.01,.02,.1,.2}$ and minimal group fraction $\in \set{.1,.2,.5}$ for (CvAR-)DORO on all datasets. 
We tune the learning rate $\in \set{10^{-5}, 10^{-4}, 10^{-3}}$ and weight decay $\in \set{10^{-4}, 10^{-3}, 10^{-2}}$ for all methods. 
We select the step size of the group weights $q$ in gDRO~\citep{sagawa2019distributionally} $\in \set{.001, .01, .1}$. 
We train a three-layer ReLU neural network with 50 hidden neurons per layer for 50 and 300 epochs on the Synthetic and COMPAS datasets, respectively.
We train a logistic regression model with 360 epochs on the Waterbirds dataset, and a BERT model for 10 epochs on the Civilcomments dataset.

\begin{figure}[h]
	\centering
	\includegraphics[width=\textwidth]{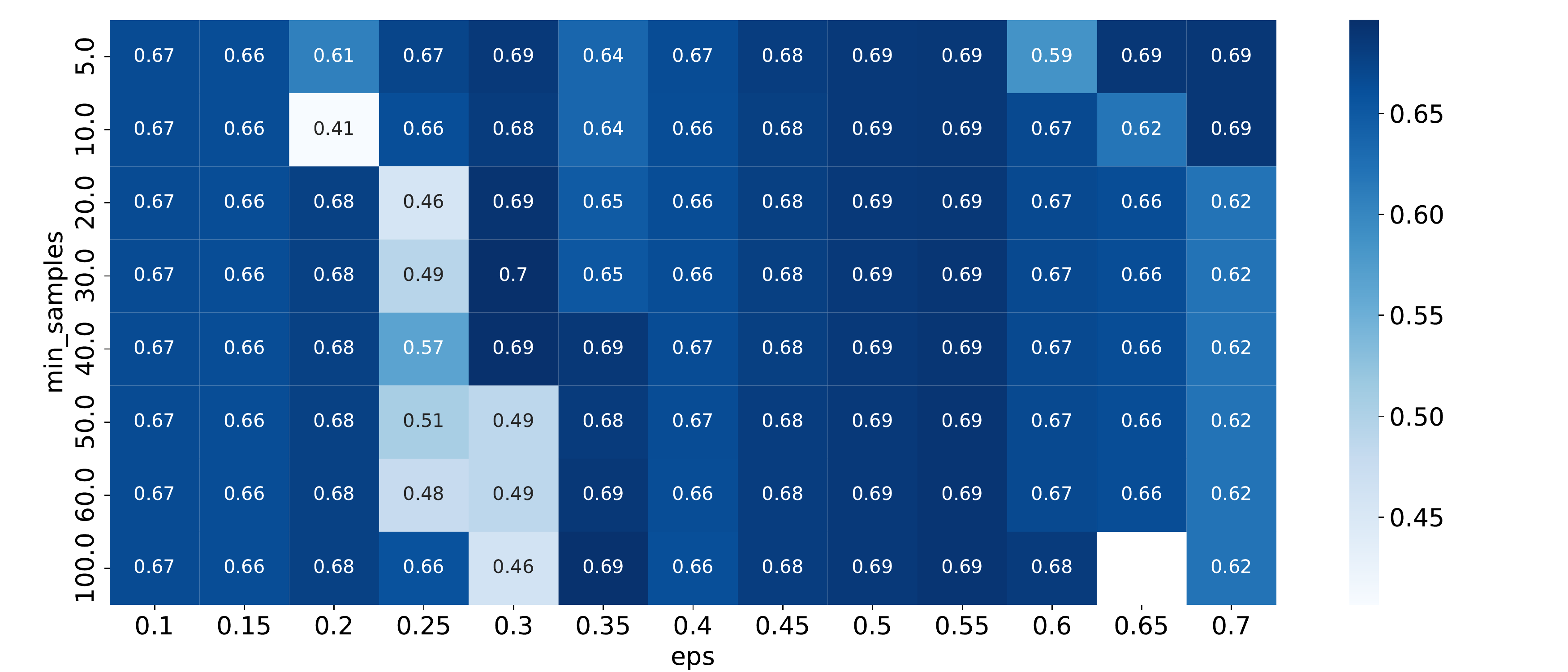}
	\caption{Worst-group accuracy of \grasp{} v.s. DBSCAN clustering hyperparameters (\texttt{eps}: $\eps$, \texttt{min\_samples}: $m$) on Waterbirds dataset.}
	\label{fig:waterbirds_heatmap_acc}
	\vspace{-.2in}
\end{figure}

\begin{figure}[t]
	\centering
	\vspace{-.2in}
	\includegraphics[width=\textwidth]{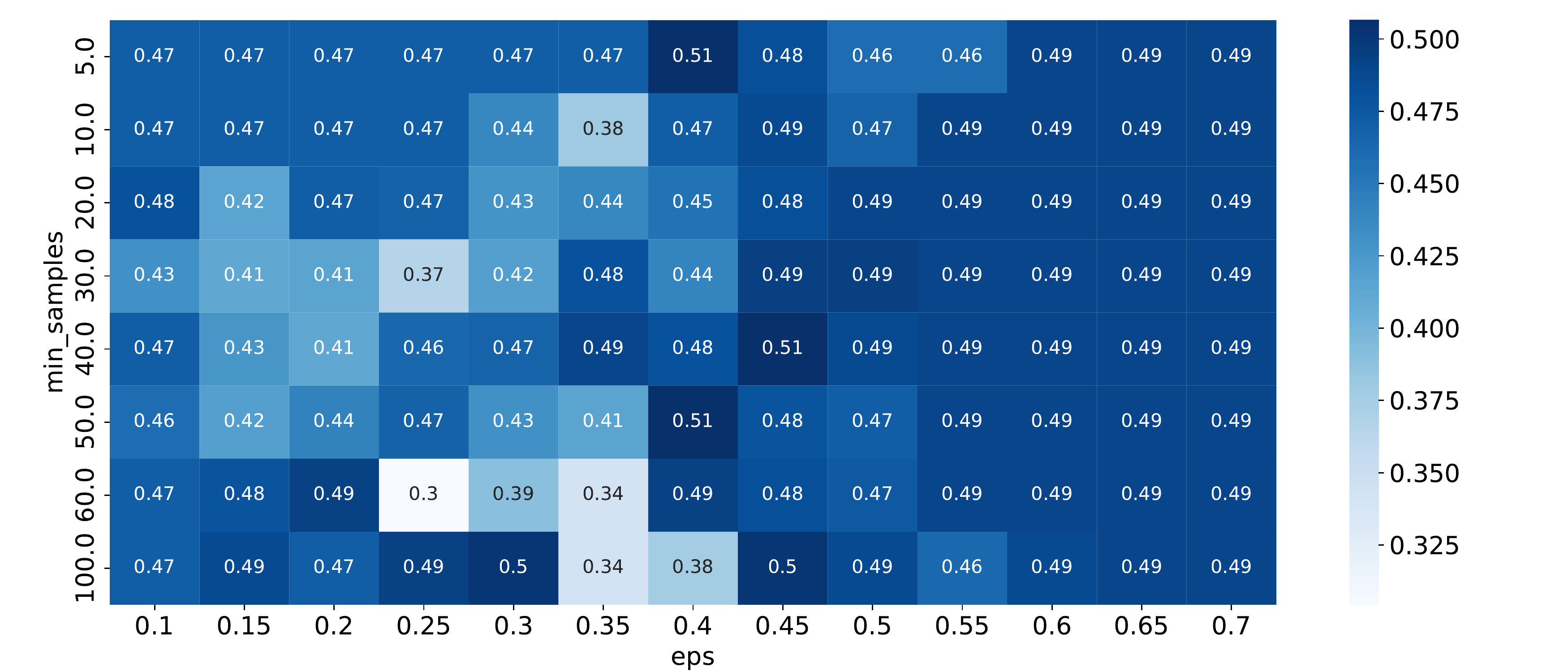}
	\caption{Worst-group accuracy of \grasp{} v.s. DBSCAN clustering hyperparameters (\texttt{eps}: $\eps$, \texttt{min\_samples}: $m$) on COMPAS dataset.}
	\label{fig:compas_heatmap_acc}
	\vspace{-.2in}
\end{figure}

\subsection{Robustness to DBSCAN Clustering Hyperparameters}\label{sec:robust_cluster}
In this experiment, we investigate the effect of clustering hyperparameters on group inference and downstream DRO task performances.
In doing so, we let $\eps \in \set{.1,.15,.2,.25,.3,.35,.4,.45,.5,.55,.6,.65,.7}$ and $m \in \set{5,10,20,30,40,50,60,100}$ and visualize how ARI varies with different choice of clustering hyperparameters on different classes of Waterbirds dataset in Fig.~\ref{fig:waterbirds_heatmap_ari_0} and Fig.~\ref{fig:waterbirds_heatmap_ari_1}. 
We observe that the group identification performance is robust to clustering hyperparameters. 
For worst-group performance, we set the $\eps$ and $m$ to be the same for different classes on the datasets.
We visualize how it varies with clustering hyperparameters on Waterbirds and COMPAS dataset in Fig.~\ref{fig:waterbirds_heatmap_acc} and Fig.~\ref{fig:compas_heatmap_acc}. 
A similar phenomenon is observed for worst-group performance — we find that worst-group performance is fairly robust to DBSCAN clustering hyperparameters.

% \begin{figure}
	%     \centering
	%     \includegraphics[width=\textwidth]{iclr2023/figures/civilcomment_0.png}
	%     \caption{Group identification quality of \grasp{} v.s. DBSCAN clustering hyperparameters (\texttt{eps}: $\eps$, \texttt{min\_samples}: $m$) measured in Adjusted Rand Index (ARI) on class 0 of Civilcomments dataset.}
	%     \label{fig:civilcomments_heatmap_ari_0}
	% \end{figure}

% \begin{figure}
	%     \centering
	%     \includegraphics[width=\textwidth]{iclr2023/figures/civilcomments_1.png}
	%     \caption{Group identification quality of \grasp{} v.s. DBSCAN clustering hyperparameters (\texttt{eps}: $\eps$, \texttt{min\_samples}: $m$) measured in Adjusted Rand Index (ARI) on class 1 of Civilcomments dataset.}
	%  \label{fig:civilcomments_heatmap_ari_1}
	% \end{figure}

\subsection{Results with Known Validation Group Annotations}\label{sec:w_val}

\begin{table}[h]
	\caption{
		\textbf{Group identification performance of Group-learning methods measured by Adjusted Rand Index (ARI) when validation group annotations are available. }
		Higher ARI indicated higher group identification quality. 
		The results are reported on clean and contaminated versions of Synthetic and Waterbirds datasets, COMPAS and Civilcomments datasets. 
		Our \grasp{} significantly outperforms the other Group-learning baselines on all the tested datasets.
		Moreover, we observe that \grasp{} is robust to outliers. 
		% \yuchen{FeaSP on Civilcomments has not been finished yet.}
	}
	\label{tab:ari_val}
	% \vskip 0.15in
	\begin{center}
		\begin{small}
			
			%\tiny{
				\begin{sc}
					
					\begin{tabularx}{.9\textwidth}{lcccccc}
						\toprule
						& \multicolumn{2}{c}{Synthetic} & \multicolumn{2}{c}{Waterbirds} & \multirow{2}{*}{COMPAS} &  \multirow{2}{*}{Civilcomments}\\
						Outliers? & \xmark & \cmark & \xmark & \cmark & &  \\
						Method & ARI & ARI & ARI& ARI& ARI& ARI \\
						\midrule
						EIIL & -.0069& -.0031 & .0114 & .0078 & -.0025 & -.0001  \\
						George &.6027& .4565 & .3223 & .3822 & .2059 & .2218  \\
						FeaSP & .5189 & .5276 & .5189 & .1069 & .2956 & .2072 \\ 
						\grasp{} (Ours) & \textbf{.7497} & \textbf{.7241} & \textbf{.8137} & \textbf{.7531} & \textbf{.5453} & \textbf{.2863}\\
						\bottomrule
					\end{tabularx}
				\end{sc}
				% }
		\end{small}
	\end{center}
\end{table}

\begin{table}[h]
	\caption{
		\textbf{Downstream DRO performance of various methods measured by worst-group accuracy and average accuracy on the test dataset when validation group annotations are available. }
		% {\color{red} KG: what do these arrows mean?} \yuchen{The values of the two metrics are the  higher the better :) }}
	The average test accuracy is a re-weighted average of the group-specific accuracies, where the weights are based on the training distribution.
	The results are reported on clean and contaminated version of Synthetic and Waterbirds datasets, COMPAS, and Civilcomments datasets. 
	% The gDRO (oracle) here is a Group-aware method, which can get access to the true group annotations.
	We observe that \grasp{} significantly outperforms the Group-oblivous (DORO) and Group-learning approaches (EIIL, George, FeaSP) methods on Synthetic, COMPAS, and Civilcomments datasets, and performs relatively well on Waterbirds datasets, while being robust to outliers. 
	Note that \grasp{} sometimes outperforms gDRO (oracle), which can get access to the true group annotations.
	This is because \grasp{} may focus on ``harder'' instances more, which potentially affect the results most.
	% \yuchen{How to write the excuse for this missing entry?}{\color{red} KG: I don't really think there can be an excuse that reviewers would allow here (they will say the paper is not ready). Probably best to simply leave blank without explanation, and provide when the reviewers ask}
}
\label{tab:worst_val}
% \vskip 0.15in
\begin{center}
	\begin{small}
		\tiny{
			\begin{sc}
				\setlength{\tabcolsep}{8pt}
				\begin{tabularx}{\textwidth}{lcccccc}
					\toprule
					& \multicolumn{2}{c}{Synthetic} & \multicolumn{2}{c}{Waterbirds} & \multirow{2}{*}{COMPAS} &  \multirow{2}{*}{Civilcomments}\\
					Outliers? & \xmark & \cmark & \xmark & \cmark & &  \\
					Method & Worst.(Avg.) & Worst.(Avg.) & Worst.(Avg.)& Worst.(Avg.)& Worst.(Avg.)& Worst.(Avg.) \\
					\midrule
					ERM & .6667(.8823) & .5333(.8273) & .6075(.9673) & .5249(.9621) & .4706(.6792) & .4659(.9213) \\ \midrule
					DORO & .6667(.8823) & .7333(.8332) & .6604(.9669) & .6056(.9066) & .4387(.6696) & .6056(.9066)\\ 
					EIIL & .6667(.8783) & .7333 (.8170) & .6927 (.9649) & \textbf{.7056(.9629)} & .0588 (.6046) & .6056(.9066) \\
					George &.6667(.8823) &  .7333(.8227) & .\textbf{8053}(.9511) & .6056(.9066) & .4664(.6219) & .6056(.9056)\\
					FeaSP & .6667(.8823) & \textbf{.8000(.8391)}& .1417(.9346) & .1417(.9346) & .4545(.6386) & .6056(.9066) \\ 
					\grasp{} (Ours) & \textbf{.8000}(.8926)& \textbf{.8000}(.8372) & .7274(.9541) & .6804(.8999) & \textbf{.4743}(.6681) & \textbf{.6804(.8999)} \\ \midrule
					gDRO (oracle) & .7333(.8639) & .8000(.8755) & .8665(.9272) & .8545(.9081) & .4625(.6807) & .6941(.8767) \\
					\bottomrule
				\end{tabularx}
			\end{sc}
		}
	\end{small}
\end{center}
\end{table}

Lastly, we report the experiment results on group inference and downstream DRO tasks where the hyperparameters of Group-oblivious (DORO) and Group-learning (EIIL, George, FeaSP, \grasp{}) are selected based on the corresponding metric computed with true validation group annotations. 
The same phenomenon we observe under the setting of unavailable validation group annotations also holds under this setting. 
Table~\ref{tab:ari_val} shows that \grasp{} also significantly outperforms the other Group-learning methods in terms of group learning when true validation group annotations are available. 
For downstream DRO tasks, while the worst-group accuracy of all Group-oblivious and Group-Learning methods are improved, \grasp{} still achieves the highest worst-group performance on Synthetic, COMPAS, and Civilcomments datasets and performs relatively well on Waterbirds dataset. 

\end{document}